\documentclass{article}

\usepackage[preprint]{neurips_2026}

\usepackage[utf8]{inputenc} % allow utf-8 input
\usepackage[T1]{fontenc}    % use 8-bit T1 fonts
\usepackage{hyperref}       % hyperlinks
\usepackage{url}            % simple URL typesetting
\usepackage{booktabs}       % professional-quality tables
\usepackage{amsfonts}       % blackboard math symbols
\usepackage{nicefrac}       % compact symbols for 1/2, etc.
\usepackage{microtype}      % microtypography
\usepackage{xcolor}         % colors
\usepackage{amsmath}
\usepackage{graphicx}
\usepackage{subcaption}
\usepackage{multirow}
\usepackage{wrapfig}
\usepackage{nicematrix}
\usepackage{array}

\usepackage[utf8]{inputenc} % allow utf-8 input
\usepackage[T1]{fontenc}    % use 8-bit T1 fonts

\usepackage[table]{xcolor}
\definecolor{ourrow}{HTML}{EBDFDF}

\def\our{ProDG}
\def\backbone{\Phi}

\usepackage[table]{xcolor}
\definecolor{best}{RGB}{255,220,220}

\title{\our{}: Prototypes for Data-Free Generative Post-Hoc Explainability}

\author{
  {\bf Piotr Borycki} \\
  Jagiellonian University \\
  IDEAS Research Institute\\
  \texttt{piotr.borycki@student.uj.edu.pl} \\
  \and
  {\bf Magdalena Tr\k{e}dowicz} \\
  Jagiellonian University \\
  \texttt{magdalena.tredowicz@student.uj.edu.pl} \\
  \and
  {\bf Jacek Tabor} \\
  Jagiellonian University \\
Centre for Credible AI\\
 Warsaw University of Technology\\
  \and
  {\bf {\L}ukasz Struski} \\
  Jagiellonian University \\
  \and
  {\bf Przemys{\l}aw Spurek} \\
  Jagiellonian University \\
  IDEAS Research Institute\\
}

\begin{document}

\maketitle

 \begin{figure}[h!]
 \vspace{-1.5cm}
     \centering
       \includegraphics[width=1\textwidth,trim={0 40 0 140},clip]{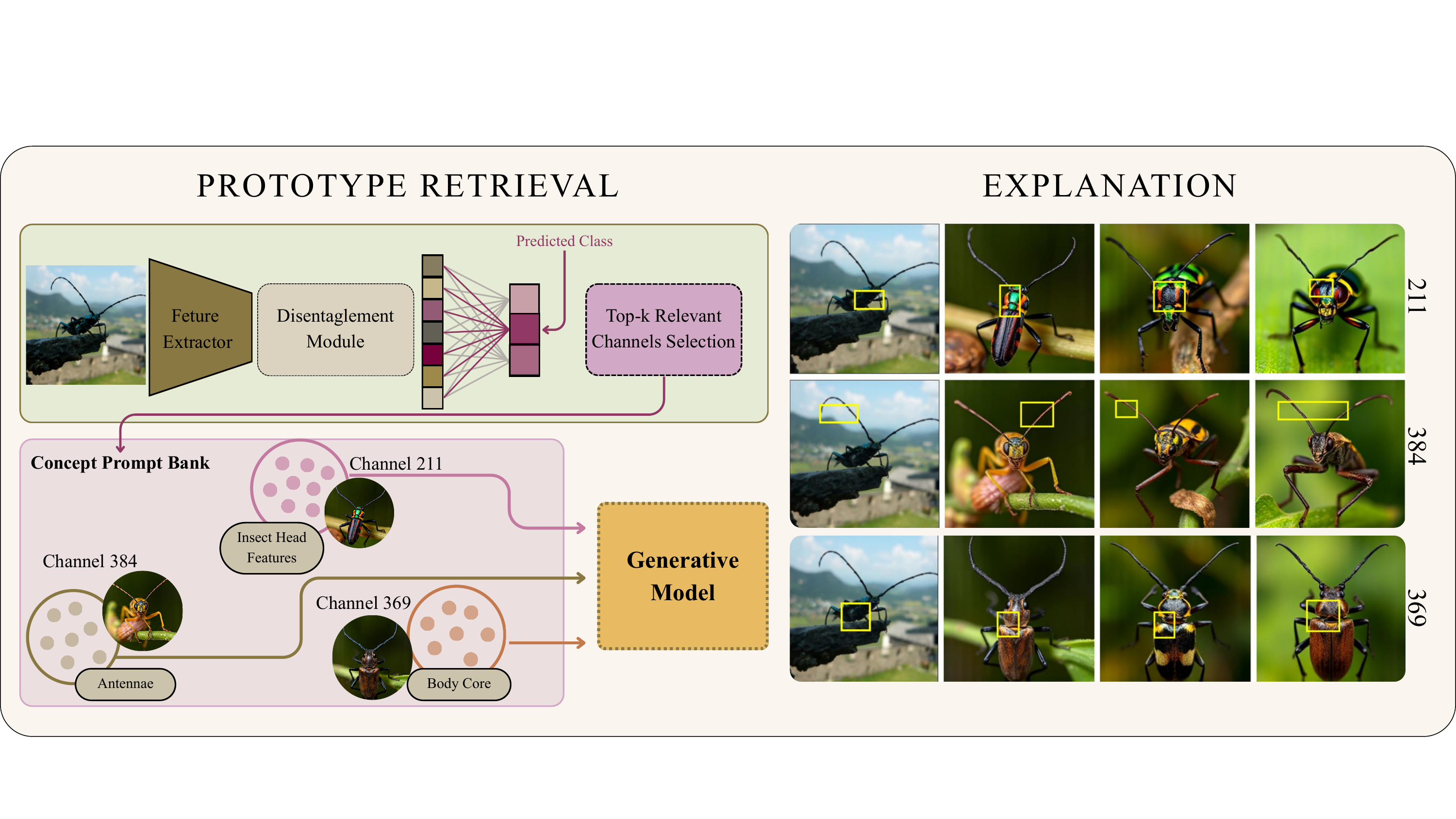}
       \vspace{-0.8cm}
     \caption{\textbf{Overview of the \our{} prototype retrieval pipeline and explanation generation.} Given an input image, \our{} first applies a classification model to predict the target class. It then identifies the top-$k$ most influential channels with respect to the predicted class. From an optimized prompt bank, \our{} samples prompts to generate corresponding images, which are subsequently used to compute activation heatmaps and derive bounding boxes for visual explanations. }
    \label{fig:teaser}
 \vspace{-0.4cm}
 \end{figure}

\begin{abstract}
  Ante-hoc interpretability methods based on prototypes provide highly accurate explanations by utilizing the intuitive "this looks like that" reasoning paradigm. On the other hand, post-hoc models can explain predictions for a single image without relying on an underlying dataset or requiring costly neural network retraining. Recent approaches successfully solve the retraining problem for prototype-based networks. However, they still face a fundamental limitation: they require access to a subset of data (e.g., a test or validation set) to search for and extract the visual prototypes. In this paper, we address this issue and introduce \our{}: Generative Prototypes for Data-Free Post-Hoc Explainability, a novel framework that leverages generative models to synthesize pure, high-fidelity prototypes directly from the frozen model's weights, completely eliminating the dependency on any external data. By establishing this new frontier in Data-Free XAI, \our{} unlocks robust visual interpretability for privacy-sensitive domains, where original data is strictly restricted or fundamentally inaccessible.  
  \\
  Project page: \url{https://github.com/piotr310100/ProDG}
\end{abstract}

\section{Introduction}

\begin{figure}[t]
    \centering

    \makebox[0.32\textwidth]{\text{\our{} (Ours)}}\hfill
    \makebox[0.32\textwidth]{\hspace{2em}\text{Grad-CAM}}\hfill
    \makebox[0.32\textwidth]{\hspace{2em}\text{LRP}}\\[4pt]

    \includegraphics[height=3.5cm, trim={18, 0, 280, 0}, clip]
    {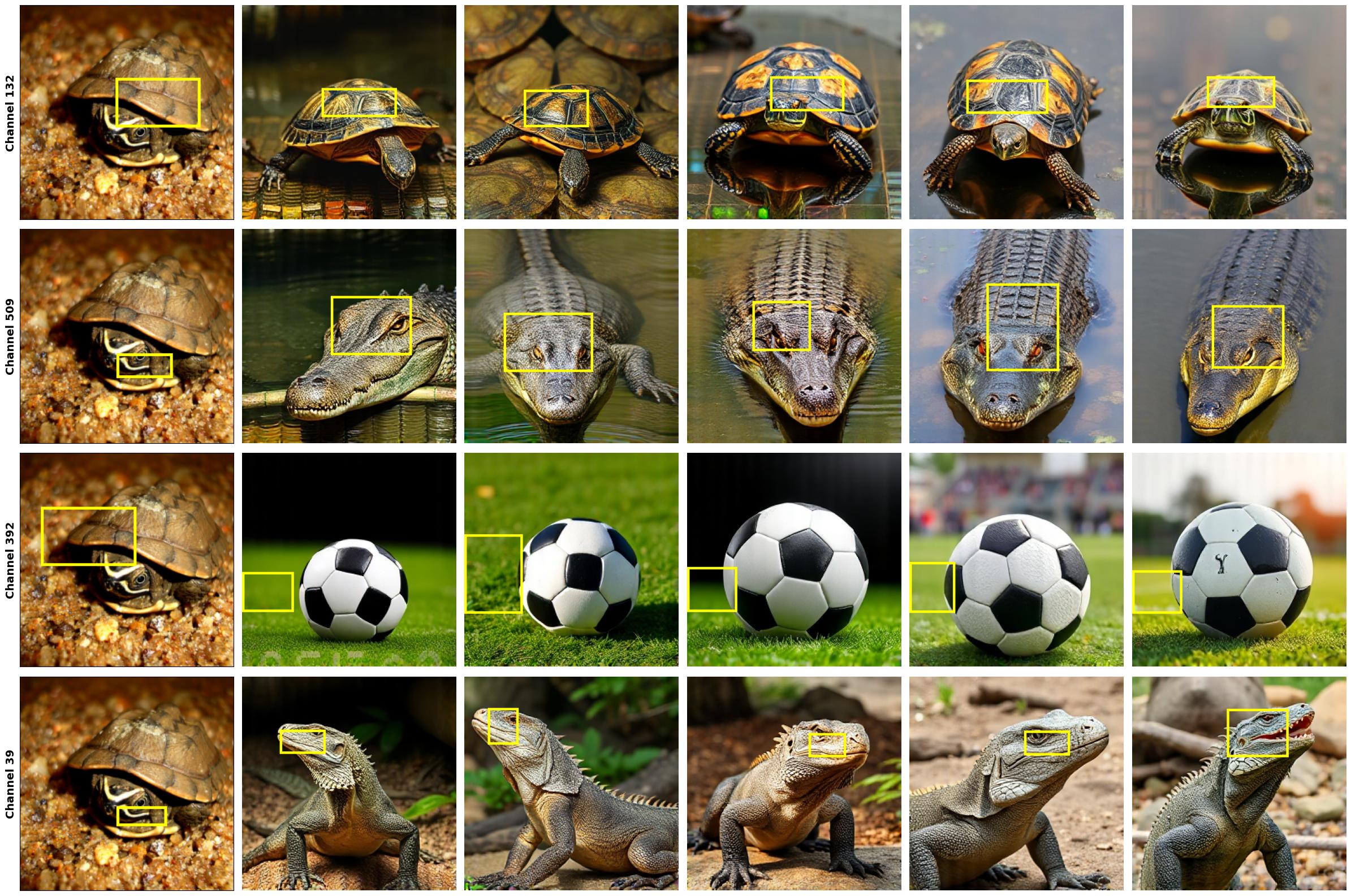}
    \hfill
    \includegraphics[height=3.5cm]
    {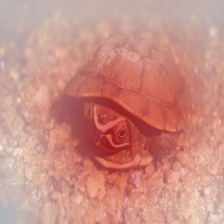}
    \hfill
    \includegraphics[height=3.5cm, trim={452, 0, 0, 0}, clip]
    {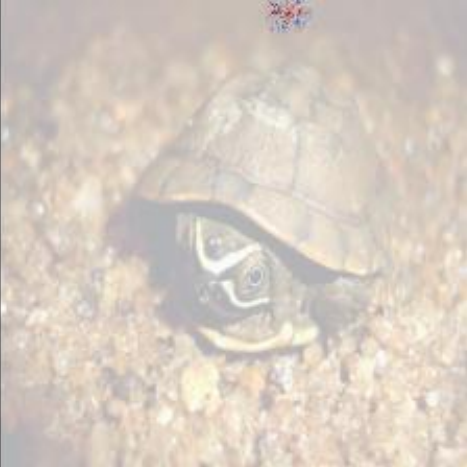}

    \caption{\textbf{Qualitative comparison of explanations produced by \our{}, Grad-CAM, and LRP} \our{} reveals semantically rich visual cues, capturing object structure, color patterns, texture details, and discriminative regions such as the tortoise shell. In contrast, Grad-CAM and LRP primarily highlight coarse activation regions, which are less informative for interpreting specific visual concepts and attributes.}
    \label{fig:comparison_gradcam_gradcam_pp}
    \vspace{-0.7cm}
\end{figure}

Deep neural networks have achieved remarkable performance in computer vision, but their black-box nature raises critical trust and safety issues. To address this, post-hoc explainability methods, such as SHAP \cite{lundberg2017unified}, LIME \cite{ribeiro2016should}, LRP \cite{bach2015pixel}, and Grad-CAM \cite{selvaraju2020grad}, were developed. These models can explain predictions for a single image without relying on an underlying dataset or requiring costly neural network retraining. While highly efficient, these standard post-hoc methods typically produce pixel-level attribution maps or feature importance scores. Unfortunately, these heatmaps often lack a clear semantic meaning, leaving the end user to guess the actual reasoning behind the model's focus. Fig~\ref{fig:comparison_gradcam_gradcam_pp} provides a qualitative comparison between \our{} prototype-based explanations and post-hoc attribution methods, including Grad-CAM and LRP.

To provide highly accurate and human-understandable explanations, ante-hoc interpretability methods based on prototypes were introduced. Pioneered by ProtoPNet \cite{chen2019looks}, these architectures utilize the intuitive "this looks like that" reasoning paradigm. The network learns a set of latent prototypes during training and classifies an image based on its visual similarity to these learned concepts. This field has seen significant development with methods improving prototype diversity, hierarchy, and mathematical guarantees, including ProtoPool \cite{rymarczyk2022interpretable}, ProtopShare \cite{rymarczyk2021protopshare}, ProtoTree \cite{nauta2021neural}, ProtKNN \cite{ukai2022looks}, and PiPNet \cite{nauta2023pip}. Furthermore, recent works like InfoDisent \cite{struski2024infodisent} and SIDE \cite{dubovik2025side} push the boundaries of prototype scalability, applying these interpretable structures to large and complex datasets. However, all these ante-hoc models inherently demand specialized architectures and extensive retraining from scratch, which is often computationally prohibitive.

Bridging the gap between post-hoc efficiency and prototype-based clarity is a significant challenge. Recent methods like EPIC~\cite{borycki2026epic} successfully solve the retraining problem for prototype-based networks. EPIC extracts prototype-like explanations from already trained, frozen classifiers in a purely post-hoc manner. Despite this breakthrough, EPIC and all previously mentioned prototype networks still face a fundamental limitation: they require access to a subset of data (e.g., a test or validation set) to search for, crop, and extract the visual representations of the prototypes. In real-world scenarios, especially in privacy-sensitive domains such as healthcare and finance, original datasets are strictly restricted or fundamentally inaccessible, rendering these data-dependent methods unusable.

In this paper, we address this critical issue and introduce \our{}: Generative Prototypes for Data-Free Post-Hoc Explainability. \our{} is a novel framework that completely eliminates dependence on external data. Instead of searching for prototype patches within a given dataset, our method leverages advanced generative models to synthesize pure, high-fidelity visual prototypes directly from the frozen model's weights. By optimizing the generative process to maximize activation purity for specific classifier channels, \our{} creates isolated, semantically meaningful visual concepts. For the overview of the proposed method see Fig.~\ref{fig:teaser}.

By establishing this new direction in Data-Free XAI, \our{} unlocks robust visual interpretability for highly constrained environments where data access is impossible. Our approach not only guarantees privacy preservation but also yields cleaner conceptual representations than traditional cropped patches, which often suffer from irrelevant background noise.

In summary, our main contributions are as follows:
\begin{itemize}
    \item We propose \our{}, a novel Data-Free Post-Hoc Explainability framework that synthesizes prototype explanations without requiring any access to the original or validation datasets.

    \item We introduce a purity-optimized conditioning mechanism that utilizes generative models to create high-fidelity, interpretable visual concepts directly from the weights of a frozen classifier.

    \item We demonstrate that generative prototypes provide superior semantic clarity and offer a secure, privacy-preserving alternative to traditional data-dependent prototype extraction.
\end{itemize}

% ================================================
% ================================================
% ================================
% ====================
% ==========
% =====
% ==
% =

\section{Related Works}

Explainability in deep learning has been approached from two fundamentally different perspectives: post-hoc methods that explain pretrained models without modifying them, and ante-hoc methods that build interpretability directly into the model architecture. Prototype-based approaches form a prominent subclass of the latter, offering human-understandable visual explanations grounded in learned concepts. Recent efforts have begun to bridge these two paradigms, though a critical gap around data accessibility remains unaddressed.

\textbf{Post-hoc explainability methods} constitute the most widely adopted family of approaches for interpreting neural network predictions. Attribution-based techniques such as LIME \cite{ribeiro2016should} and SHAP~\cite{lundberg2017unified} explain individual predictions by approximating model behavior locally or via Shapley value decomposition, respectively. Gradient-based and propagation-based methods, including LRP~\cite{bach2015pixel} and Grad-CAM~\cite{selvaraju2020grad}, generate spatial saliency maps by propagating relevance signals backward through the network or leveraging class-discriminative gradient information. While computationally efficient and architecture-agnostic, these methods produce pixel-level heatmaps that frequently lack human-interpretable semantic meaning, requiring the user to infer the conceptual rationale behind a model's focus.

\textbf{Prototype-based interpretable architectures} address this semantic gap by grounding classification decisions in human-understandable visual concepts. ProtoPNet~\cite{chen2019looks} pioneered this paradigm by learning a set of latent prototypical parts during training and classifying images through their similarity to these prototypes, embodying the "this looks like that" reasoning principle. Subsequent work has substantially extended this framework along multiple dimensions. ProtoPool~\cite{rymarczyk2022interpretable} introduces a differentiable prototype assignment mechanism, enabling more flexible prototype allocation across classes. ProtopShare~\cite{rymarczyk2021protopshare} explores cross-class prototype sharing to reveal structural similarities between categories. ProtoTree~\cite{nauta2021neural} organizes prototypes hierarchically in a decision tree, yielding more compositional and interpretable reasoning paths. PiPNet~\cite{nauta2023pip} shifts the focus to patch-based prototypes with stronger alignment guarantees, improving visual coherence and faithfulness. ProtKNN~\cite{ukai2022looks} reformulates prototype-based reasoning within a similarity-based nearest-neighbor classification framework. Collectively, these methods demonstrate the richness and versatility of the prototype paradigm, yet they universally require purpose-built architectures trained from scratch, imposing substantial computational costs.

\textbf{Bridging post-hoc efficiency and prototype-based clarity} remains an active research direction. EPIC~\cite{borycki2026epic} makes a significant contribution by extracting prototype-like explanations from already-trained, frozen classifiers in a purely post-hoc manner, eliminating the need for specialized architecture design or retraining. Similarly, InfoDisent~\cite{struski2024infodisent} advances the scalability of interpretable representations by disentangling information in the latent space of deep networks, enabling richer explanations on large and complex datasets. Despite these advances, a critical limitation persists across the entire landscape of prototype-based methods: they all require access to an external dataset -- typically a validation or test set -- to search for, crop, and visualize prototype representatives. This dependency renders such methods inapplicable in privacy-constrained environments, such as healthcare, where access to original training data is strictly prohibited. Our proposed method addresses precisely this gap by generating synthetic prototypes directly from the weights of a frozen model, without requiring any data access.

% ================================================
% ================================================
% ================================
% ====================
% ==========
% =====
% ==
% =

\section{Method}
\begin{figure}[t]
    \centering
    \includegraphics[width=\linewidth]{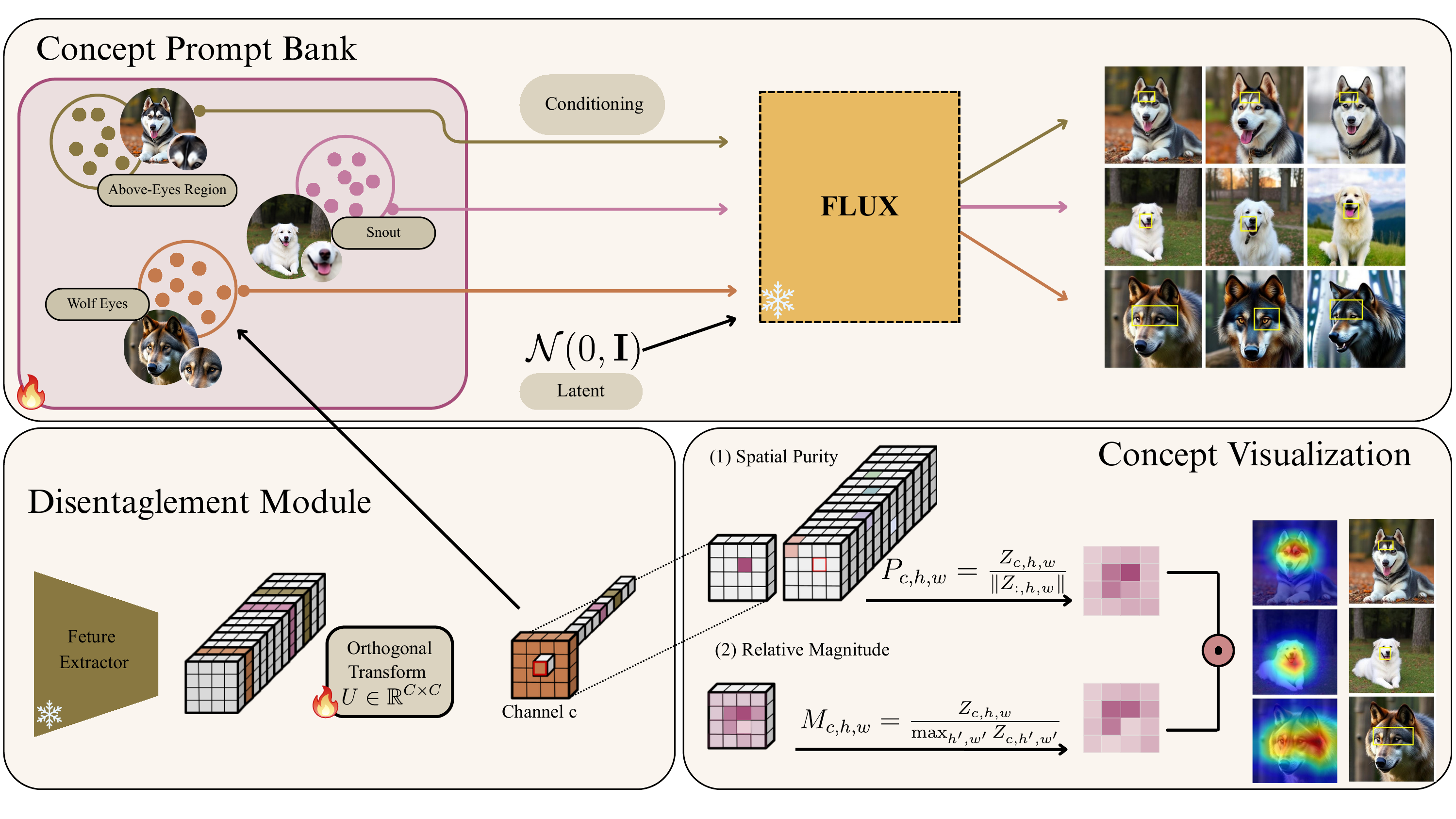}
    \caption{\textbf{Our framework \our{} performs concept prompts optimization and feature disentanglement within the FLUX generative model.} The Concept Prompt Bank parameterizes the text embeddings of a frozen generative model (FLUX) to synthesize prototypical images that maximize concept purity. This prompt bank uses a reparameterization trick offsets to ensure a diverse distribution of generated images. The Orthogonal Feature Disentanglement Module learns an orthogonal transformation $U \in \mathbb{R}^{C \times C}$ that maps entangled features from a frozen backbone into a space where individual channels $c$ correspond to unique visual concept. During Concept Visualization, the framework identifies prototypical regions by calculating the pixel-wise product of Spatial Purity $P_c$ and Relative Magnitude $M_c$, resulting in heatmaps and localized bounding boxes.}
    \label{fig:architecture}
    \vspace{-0.5cm}
\end{figure}

The main goal of this work is to extract human-interpretable visual concepts, including shapes and object parts, from the pretrained image classification network feature space. Recent approaches, such as EPIC~\cite{borycki2026epic}, successfully managed to find these concepts and represent them using visual prototypes. However, they require access to a large real-world dataset that the original classifier was trained on. This limits the discoverable concepts to the specific distribution of the provided dataset, limiting their applicability when the original data is unavailable.

To address this, we propose a novel data-free framework, that takes advantage of the visual knowledge of a trained generative model to discover concept prototypes. Instead of searching for concepts within a fixed dataset, our approach optimizes the generator's embedding space. The goal is to find embeddings that generate images, which isolate and maximize specific concepts within the classifier without ever accessing real data. To the best of our knowledge, this is the first prototype-based method, which does not require dataset access for visual concepts discovery.

The proposed solution consists of three main components: (1) an Orthogonal Feature Disentanglement Module that transform the classifier's feature space to separate overlapping representations, (2) a Concept Prompt Bank that parameterizes the text embeddings conditioning the generative model, and (3) an alternating optimization scheme that maximizes the concept's purity while ensuring diversity. Fig.~\ref{fig:architecture} presents our framework and interactions between main components.

\paragraph{Orthogonal Feature Disentanglement} Consider a pretrained image classification network where $\backbone: \mathcal{X} \to \mathbb{R}^{C \times H \times W}$ represents its feature extractor. In standard networks, a single visual feature is typically entangled and distributed across multiple channels. To isolate these features, we introduce an orthogonal change of basis matrix $U \in \mathbb{R}^{C \times C}$ that maps the original entangled activations into a new space $Z = U\, \backbone(x)$. The intuition is that applying an orthogonal change of basis to the feature space can align the new channel representations with distinct, semantically meaningful visual concepts, ensuring that each channel $c \in \{1, \dots, C\}$ activates strongly for exactly one specific feature.

To quantify this isolation, we evaluate the activation strength of generated images using the \textit{purity} metric, inspired by EPIC~\cite{borycki2026epic}. For a given input image $x$ and target channel $c$, let $(h, w)$ be the spatial coordinates of the maximum activation in the transformed channel $Z_{c} = (U \, \backbone(x))_c$. The purity is defined as the ratio of this maximum activation to the norm of the feature vector spanning channels at that specific spatial location:
\[
    purity(x; U, c) = \frac{Z_{c, h, w}}{\| Z_{:, h, w} \|}.
\]
By maximizing purity, the method enforces a single dominant concept in the feature representation at the relevant spatial location. 

Crucially, this feature disentanglement process does not degrade the original model's predictive performance. Because $U$ is constrained to be an orthogonal matrix, its inverse can be quickly calculated using transposition ($U^{-1} = U^T$). By modifying the original linear classification head to incorporate this inverse mapping, by multiplying the classifier weights by $U^T$, the change of basis is mathematically canceled out in the final classification layer. This guarantees that the final logit predictions, and consequently the accuracy of the modified network, remain identical to those of the base model, ensuring faithfulness to the original decision process.

\paragraph{Generative Concept Discovery}
Our approach synthesizes prototype images representing each concept $c$. Let $G(\mathbf{e}_c)$ be a frozen generative model conditioned on text embeddings $\mathbf{e}_c$. Since the parameters of $G$ are frozen, our goal is to learn the optimal embeddings that force $G$ to generate images maximizing the purity of channel $c$. In this work, we focus on FLUX.1-schnell~\cite{flux2024} as the generative prior, which relies on two separate text encoders: a high-dimensional T5 encoder and a more compact CLIP encoder.

To achieve this, we introduce a \textit{Concept Prompt Bank}, denoted as $\mathcal{P}$, which is formally defined as a set of parameter tuples for all $C$ channels $\mathcal{P} = \{ \mathcal{P}_c \}_{c=1}^C$. For each channel $c$, the element $\mathcal{P}_c$ encapsulates both the frozen anchors and the learnable distributional parameters governing the embeddings. Specifically, we define $\mathcal{P}_c = (\mathbf{pe}_{\text{anchor}, c}, \mathbf{ppe}_{\text{anchor}, c}, \Theta_c)$. Here, $\mathbf{pe}_{\text{anchor}, c}$ and $\mathbf{ppe}_{\text{anchor}, c}$ are the frozen anchor embeddings corresponding to the T5 and CLIP encoders, respectively. 

To initialize these anchors with semantically meaningful prompts that generate images understood by the classifier, we perform a discovery phase. First, we generate a batch of images using the text names of the classes the classifier is capable of predicting. We then extract features for these images and calculate their purities, notably the matrix $U=\text{Id}$ at this stage. Each channel $c$ is assigned the text embedding corresponding to the class name that achieves the highest purity score for that specific channel.

The term $\Theta_c$ represents the learnable offset and variance parameters optimized during training. Due to high dimensionality of T5 embeddings, which consists of 512 tokens and embedding dimension of 4096, we parameterize it using Low-Rank Adaptation (LoRA) in the embedding dimension $\Delta \mathbf{pe}_c = A_c B_c$,
where $A_c \in \mathbb{R}^{512 \times r}$ and $B_c \in \mathbb{R}^{r \times 4096}$. For the CLIP based pooled embeddings, which are significantly smaller, we learn a direct offset $\Delta \mathbf{ppe}_c$.

To encourage the model to generate a diverse distribution of prototype images rather than a single static prototype per concept, we formulate the embeddings as a probabilistic distribution and sample from it using the reparameterization trick. Alongside the mean offsets, $\Theta_c$ includes log-variance parameters ($\log \sigma^2_{\mathbf{pe}, c}$ and $\log \sigma^2_{\mathbf{ppe}, c}$). The final embeddings fed to the generative model are computed as:
\begin{align*}
    &\mathbf{pe}_c = \mathbf{pe}_{\text{anchor}, c} + \Delta \mathbf{pe}_c + \exp\left(\frac{1}{2} \log \sigma^2_{\mathbf{pe}, c}\right) \odot \epsilon_1
    \\
    &\mathbf{ppe}_c = \mathbf{ppe}_{\text{anchor}, c} + \Delta \mathbf{ppe}_c + \exp\left(\frac{1}{2} \log \sigma^2_{\mathbf{ppe}, c}\right) \odot \epsilon_2
\end{align*}
where $\epsilon_1, \epsilon_2 \sim \mathcal{N}(0, \mathbf{I})$ are standard Gaussian noise vectors. These values can be used to generate image corresponding to selected channels. Given the sampled embeddings $\mathbf{pe}_c$ and $\mathbf{ppe}_c$, we initialize random latents and obtain the generated image $\hat{x}_c = G(\mathbf{pe}_c, \mathbf{ppe}_c)$. 

\paragraph{Optimization} Our method jointly optimizes the transformation matrix $U$ and the learnable parameters $\Theta_c$ within the Concept Prompt Bank $\mathcal{P}$. We parameterize $U$ as the matrix exponential of an anti-symmetric matrix, $U = \exp(A - A^T)$, guaranteeing strict orthogonality during optimization. 
Because modifying the prompts embeddings changes the generated images, we use an alternating optimization strategy consisting of two phases. Each batch processes $B$ generated images. To enable the pairwise diversity loss, we sample $B/2$ unique channels and generate $K=2$ images per channel.

In the first phase we freeze the parameters of the prompt bank, sample images $\hat{x}_c$ for randomly selected target channels, and update $U$ to maximize the average purity of these images:
\[
\mathcal{L}_U = - \frac{1}{B} \sum_{b=1}^B purity(\hat{x}_{c_b}; U, c_b).
\]

In the second phase we freeze $U$, generate images, and compute the purity score to update the embeddings similarly to the first phase. However, to prevent the prompts from collapsing we apply two additional losses. The first regularization loss ($\mathcal{L}_{\text{reg}}$) penalizes the norm of the learned prompt deltas $\Delta \mathbf{pe}_{c_b}$ and $\Delta \mathbf{ppe}_{c_b}$ to keep the embeddings reasonably close to the meaningful natural language anchors:
\[
\mathcal{L}_{\text{reg}} = \frac{1}{B} \sum_{b=1}^B \left( \text{MSE}(\Delta \mathbf{pe}_{c_b}) + \text{MSE}(\Delta \mathbf{ppe}_{c_b}) \right).
\]
On the other hand, diversity loss ($\mathcal{L}_{\text{div}}$) encourages the generative model to create different images each time for the same channel, consequently preventing the distribution from collapse into a single point. To achieve this we extract the classification features for the generated images, apply spatial average pooling to obtain feature vectors $\mathbf{v}_{c_b, k} = \text{AvgPool}(\backbone(\hat{x}_{c_b, k}))$, and compute the average pairwise cosine similarity among the $K$ variations for each concept:
\[
    \mathcal{L}_{\text{div}} = \frac{1}{B/2} \sum_{b=1}^{B/2}  \frac{\mathbf{v}_{c_b, 1}^T \mathbf{v}_{c_b, 2}}{\|\mathbf{v}_{c_b, 1}\| \|\mathbf{v}_{c_b, 2}\|}.
\]

The combined objective for the prompt bank is:
\(
\mathcal{L} = -\mathcal{L}_U + \lambda_{\text{reg}} \mathcal{L}_{\text{reg}} + \lambda_{\text{div}} \mathcal{L}_{\text{div}}
\)
where $\lambda_{\text{reg}}$ and $\lambda_{\text{div}}$ are hyperparameters. 
% In our experiments $\lambda_{\text{reg}}=0.5$ and $\lambda_{\text{div}}=0.1$.

\paragraph{Inference and Explanation Generation} 
After the orthogonal matrix $U$ and Concept Prompt Bank are optimized, our framework can extract human-interpretable explanations for any input image $x$ without additional training. The inference pipeline starts with concept attribution. To identify which visual concepts contribute most to the final prediction, we pass the input image $x$ through the frozen backbone and apply the learned orthogonal matrix to extract the disentangled representations $Z = U\, \backbone(x)$. Letting $W$ denote the weights of the final linear classification layer, we calculate the importance score of a specific channel $c$ towards predicted class $\hat{y}$ as
$
    S_{\hat{y},c} = W_{\hat{y},c} \cdot \text{ReLU}(\text{GAP}(Z_c)).
$
The most important concepts are selected by choosing channels corresponding to $k$ largest scores.

Next, to visualize the selected concepts, we sample from the optimized prompt bank, generate the corresponding images, and compute the activation heatmaps corresponding to $U \backbone(\hat{x}_c)$. This allows dynamic extraction of bounding boxes denoting the active prototypical regions without requiring external real-world data. To specify heatmaps, we calculate a pixel-wise product of two spatial maps for a given channel $c$: Spatial Purity ($P_c$) and Relative Magnitude ($M_c$). Let $Z^+ = \text{ReLU}(U \backbone(\hat{x}_c))$ be the transformed feature map. The spatial purity map evaluates the dominance of channel $c$ at every spatial location $(h, w)$ and is equal to
$
    P_{c, h, w} = \frac{Z^+_{c, h, w}}{\| Z_{:, h, w} \|}.
$
The relative magnitude map normalizes the channel's activation against the maximum achieved along the feature channel map
$
    M_{c, h, w} = \frac{Z^+_{c, h, w}}{\max_{h', w'} Z^+_{c, h', w'}}.
$
The final concept heatmap is given by $H_c = P_c \odot M_c$, which is then interpolated to the original image resolution. To localize the concept, we extract a bounding box from $H_c$. It is based on a binary mask created by thresholding the heatmap at $80\%$. We then look for the largest contiguous block of active neighboring pixels. The minimum and maximum spatial coordinates of this largest block define the bounding box, providing a clean, localized visual explanation of the concept.

% ================================================
% ================================================
% ================================
% ====================
% ==========
% =====
% ==
% =

\begin{figure}[t]
    \centering
    \setlength{\tabcolsep}{0.5pt}
    {
        \begin{NiceTabular}{
        % @{\extracolsep{\fill}}
        X[c] X[c] X[c]
        }[width=\textwidth]
                \Block[]{2-1}{} 
                \our{} (Ours) & 
                \Block[]{2-1}{} 
                EPIC &
                \Block[]{2-1}{} 
                InfoDisent \\
                \includegraphics[height=3.5cm, trim={18, 0, 280, 0}, clip]{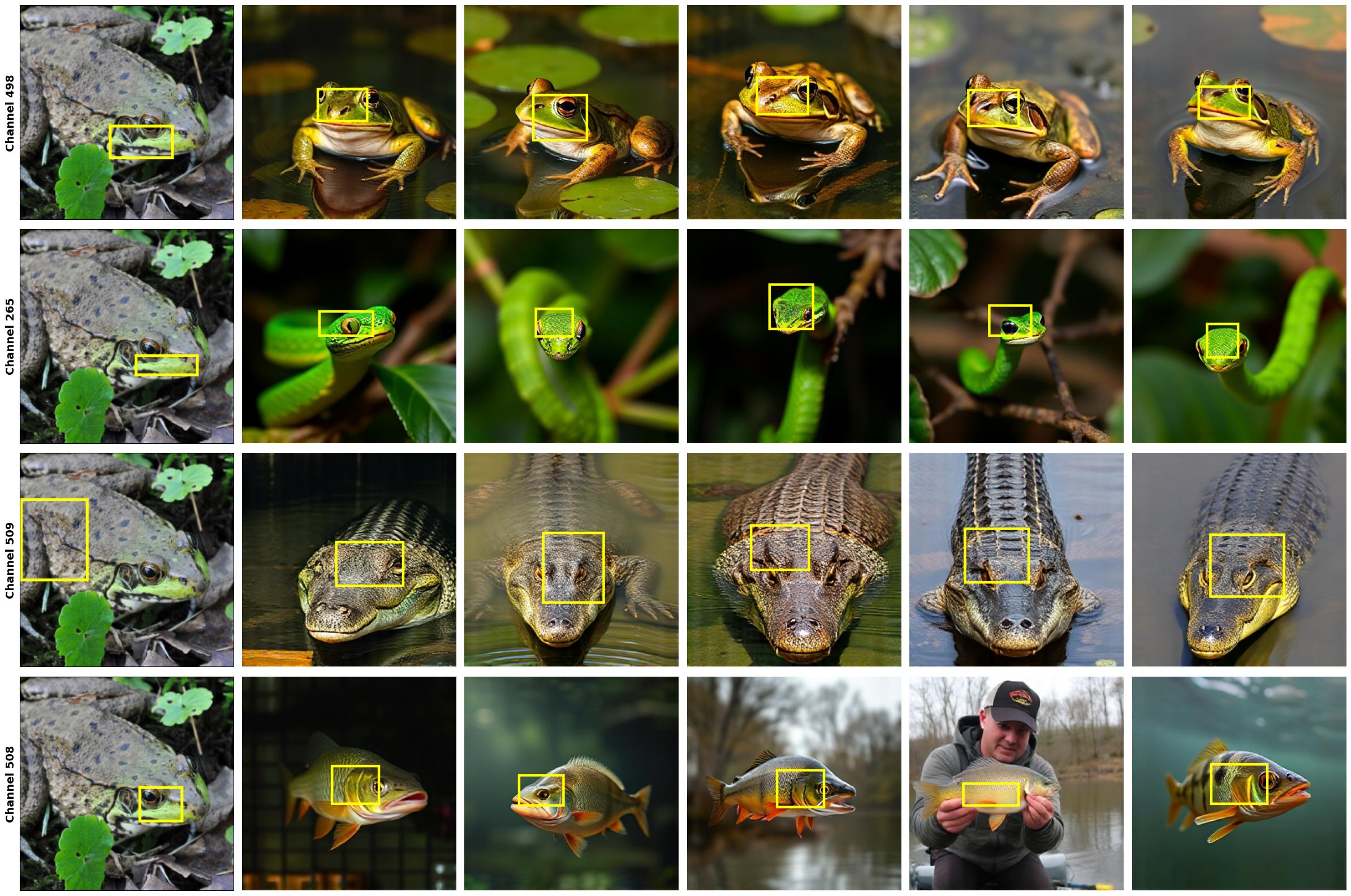} &
                \includegraphics[height=3.5cm,]{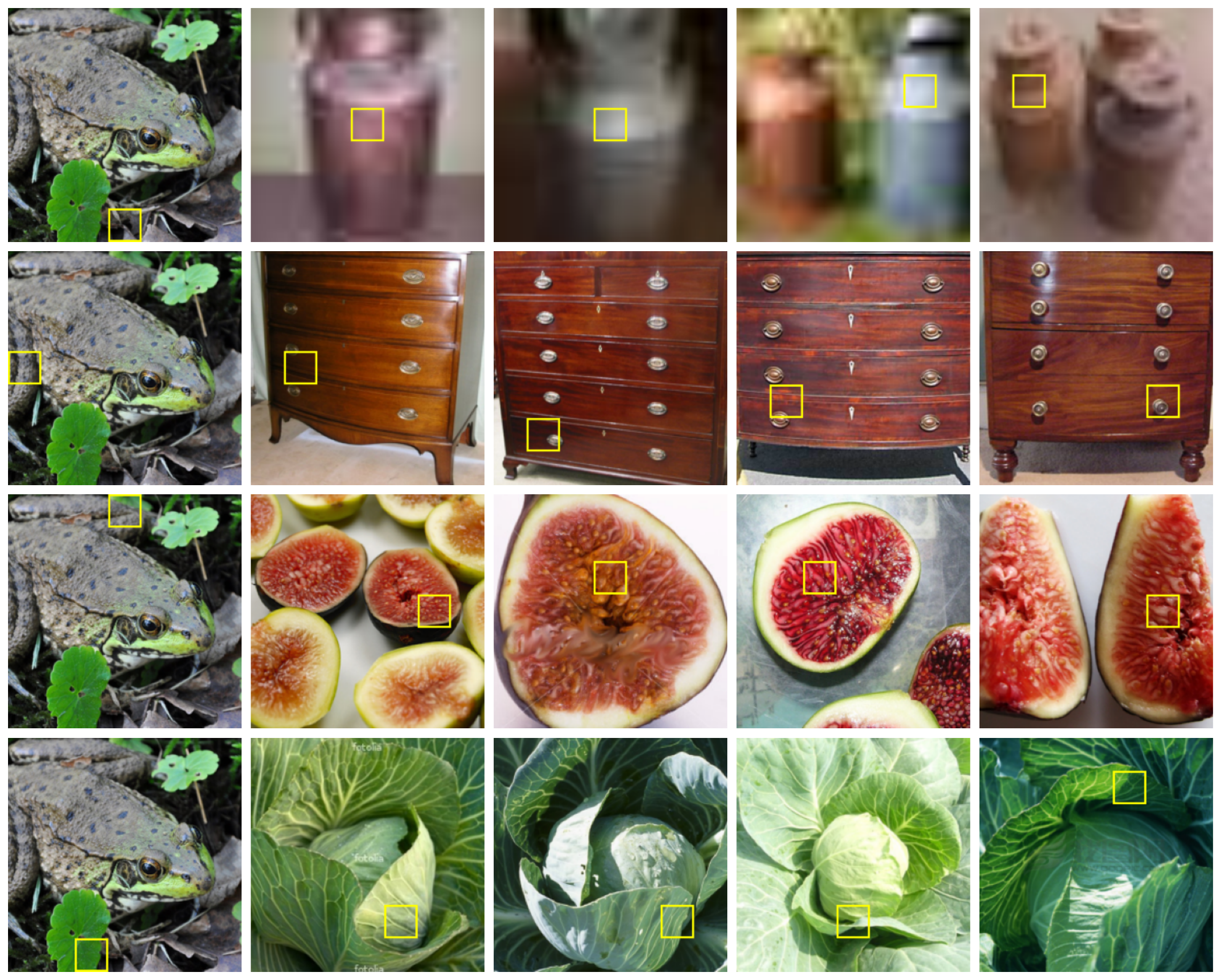} &
                \includegraphics[height=3.5cm, trim={0, 0, 225, 0}, clip]{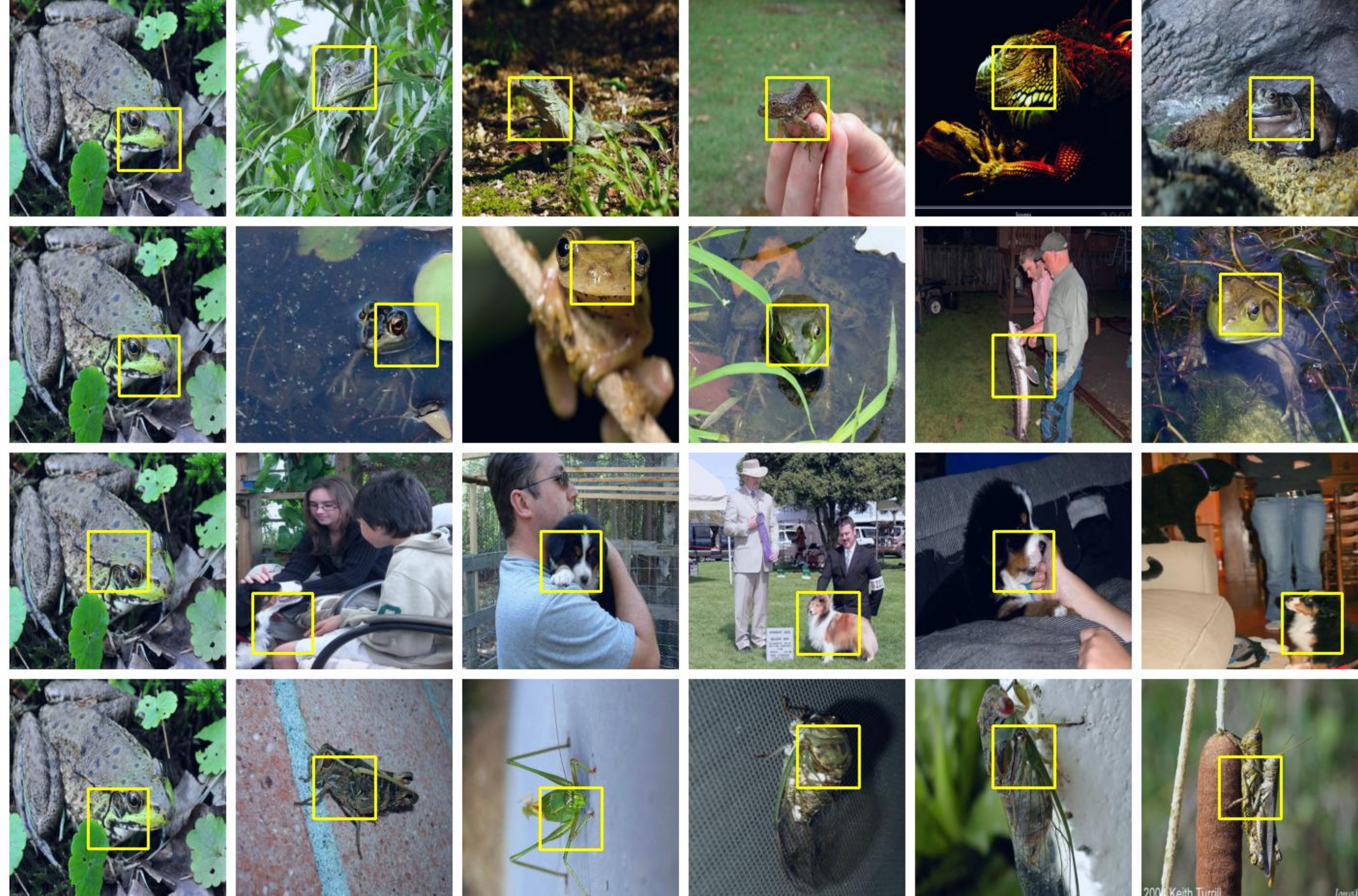}
        \end{NiceTabular}
    }
    \caption{\textbf{Comparison of explanations between \our{} (Ours) EPIC and InfoDisent.} This comparison highlights the data-independent nature of \our{}, in contrast to the data-dependent explanation mechanisms of EPIC and InfoDisent. The comparison is conducted on a representation learned on top of pretrained ResNet34.
    }    \label{fig:comparison_epic_infodisent_freegen_imagenet}
    \vspace{-0.6cm}
\end{figure}

\section{Experiments and Results}

\begin{table}[t]
  \caption{\textbf{Comparison of accuracy across datasets and backbones.} Similar to EPIC, \our{} maintains the baseline model's accuracy while matching other standard methods.}
  \label{tab:accuracy}
  \centering
  \small
  \setlength{\tabcolsep}{0pt}
  \begin{NiceTabular*}{\textwidth}{l@{\extracolsep{\fill}} l ccccc}
    \toprule
    \textbf{Dataset} & \textbf{Model} & \textbf{ResNet-34} & \textbf{ResNet-50} & \textbf{DenseNet-121} & \textbf{ConvNeXt-L} & \textbf{Swin-S} \\
    \midrule
    \Block{4-1}{ImageNet} 
    & Baseline & 73.3\% & 80.8\% & 74.4\% & 84.1\% & 83.7\% \\
    & InfoDisent & 64.1\% & 67.8\% & 66.6\% & 82.8\% & 81.4\% \\
    & EPIC & \textbf{73.3\%} & \textbf{80.8\%} & \textbf{74.4\%} & \textbf{84.1\%} & \textbf{83.7\%} \\
    % \rowcolor{ourrow}
    & \our{\,(ours)} & \textbf{73.3\%} & \textbf{80.8\%} & \textbf{74.4\%} & \textbf{84.1\%} & \textbf{83.7\%} \\
    
    \midrule
    
    \Block{5-1}{CUB-200} 
    & Baseline & 76.0\% & 78.7\% & 78.2\% & --- & --- \\
    & InfoDisent & \textbf{78.3\%} & 79.5\% & 80.6\% & --- & --- \\
    & EPIC & 76.0\% & 78.7\% & 78.2\% & --- & --- \\
    % & ProtoPNet & 74.1\% & 84.8\% & 76.6\% & --- & --- \\
    & ST-ProtoPNet & 78.2\% & \textbf{88.0\%} & \textbf{81.8\%} & --- & --- \\
    % & TesNet & 76.5\% & 87.3\% & 80.9\% & --- & --- \\
    & \our{\,(ours)} & 76.0\% & 78.7\% & 78.2\% & --- & --- \\

    \midrule

    \Block{5-1}{Dogs} 
    & Baseline & 84.5\% & 87.4\% & 84.1\% & --- & --- \\
    & InfoDisent & 83.9\% & 86.6\% & 83.8\% & --- & --- \\
    & EPIC & \textbf{84.5\%} & \textbf{87.4\%} & \textbf{84.1\%} & --- & --- \\
    % & ProtoPNet & 76.1\% & 78.1\% & 75.4\% & --- & --- \\
    & ST-ProtoPNet & 83.4\% & 83.3\% & 82.9\% & --- & --- \\
    % & TesNet & 81.2\% & 85.7\% & 82.1\% & --- & --- \\
    & \our{\,(ours)} & \textbf{84.5\%} & \textbf{87.4\%} & \textbf{84.1\%} & --- & --- \\
    \bottomrule
  \end{NiceTabular*}
  \vspace{-0.5cm}
\end{table}

In this section, we evaluate the effectiveness of our data-free concept discovery framework. The primary objective of these experiments is to demonstrate that high-quality, human-interpretable visual concepts can be extracted from a classifier's feature space using only a generative prior, without relying on the original training distribution. To validate our approach, the evaluation focuses on four key aspects: (1) the predictive faithfulness of the disentangled backbone, (2) the quantitative and qualitative performance compared to data-dependent baselines, (3) the intra-concept diversity of generated prototypes, and (4) human-centric evaluations of visual coherence and interpretability.

\paragraph{Experimental Setup} We evaluate our data-free concept discovery framework across four standard image classification benchmarks: CUB-200-2011, Stanford Cars, Stanford Dogs, and ImageNet. For the classifier backbones, we employ standard pretrained architectures, such as ResNet; DenseNet; ConvNeXT; and SwinTransformer, to demonstrate the architecture-agnostic nature of our feature disentanglement. As the generative prior, we utilize the frozen FLUX.1-schnell model. The experiments were conducted on a single NVIDIA GH200 96GB Superchip.

During the alternating optimization phase, we use the Adam optimizer. We configure the Concept Prompt Bank with a LoRA rank of $r=128$ for the T5 embeddings, while learning direct offsets for the CLIP embeddings. We set the batch size of generated images to $B=16$, sampling unique channels per iteration and generating $K=2$ variations per channel to compute the pairwise diversity loss. The objective hyperparameters are set to $\lambda_{\text{reg}}=0.5$ and $\lambda_{\text{div}}=0.1$, running for $15,000$ iterations with $1,500$ warmup steps, during which only the orthogonal matrix $U$ was optimized. We compare our method against two primary baselines: EPIC~\cite{borycki2026epic} and InfoDisent~\cite{struski2024infodisent}. 

% The concept discovery phase, which optimizes the prompt bank and orthogonal matrix, takes approximately 10 to 15 hours for the largest classification backbones on a single NVIDIA GH200 GPU. However, it is important to emphasize that this is a strictly one-time procedure per frozen classifier. Once the prompt bank is optimized, the inference phase for extracting explanations on novel images is highly efficient, requiring only a single forward pass through the classifier and one image generation step.

\paragraph{Quantitative and Qualitative Comparison}

One of the requirements of post-hoc methods is the faithfulness to the original model. The formulation of \our{} guarantees the predictive performance is not altered by the change of basis. This is due to the matrix $U$ being constrained to be orthogonal and integrating its inverse inverse into the weights of the classifier's linear head, which cancels out during the forward pass. Consequently, the accuracy difference between the original entangled model and our disentangled model is identically zero across all datasets. To validate this claim experimentally and compare the results with existing methods, we provide results on multiple datasets and classifiers in Tab.~\ref{tab:accuracy}.

Furthermore, when compared to EPIC and InfoDisent, our data-free approach yields highly competitive visual concepts, see Fig.~\ref{fig:comparison_epic_infodisent_freegen_imagenet}. Our prompt bank successfully parametrizes the generative space to synthesize comparable prototypical features, demonstrating that access to the original data distribution is not strictly necessary for high-quality concept discovery. Additional results can be found in Appendix~\ref{app:comparisons}.

% \begin{figure}[t]
%     \centering

%     \makebox[0.32\textwidth]{\text{\our{} (Ours)}}\hfill
%     \makebox[0.32\textwidth]{\hspace{2em}\text{Grad-CAM}}\hfill
%     \makebox[0.32\textwidth]{\hspace{2em}\text{Grad-CAM++}}\\[4pt]

%     \includegraphics[height=3.5cm, trim={18, 0, 280, 0}, clip]
%     {images/comparison_gradcam_gradcam_pp/explanation_1800.jpg}
%     \hfill
%     \includegraphics[height=3.5cm]
%     {images/comparison_gradcam_gradcam_pp/gradcam_1800.png}
%     \hfill
%     \includegraphics[height=3.5cm]
%     {images/comparison_gradcam_gradcam_pp/gradcam_pp_1800.png}

%     \caption{\textbf{Qualitative comparison of explanations produced by \our{}, Grad-CAM, and Grad-CAM++.} \our{} reveals semantically rich visual cues, capturing object structure, color patterns, texture details, and discriminative regions such as the tortoise shell. In contrast, Grad-CAM and Grad-CAM++ primarily highlight coarse activation regions, which are less informative for interpreting specific visual concepts and attributes.}
%     \label{fig:comparison_gradcam_gradcam_pp}
%     \vspace{-0.6cm}
% \end{figure}

\paragraph{Intra-Concept Diversity}

\begin{table}[t]
\caption{\textbf{Intra-concept diversity of generated prototypes evaluated using pairwise LPIPS.} Higher values indicate greater perceptual variance among images generated for the same concept. The consistently high scores ($\ge 0.61$) across multiple architectures demonstrate that our framework effectively avoids mode collapse, producing diverse visual explanations.}
% \vspace{0.1cm}
\label{tab:combined_lpips}
\centering
\small
% \fontsize{6.8pt}{11pt}\selectfont{
\begin{tabular}{lccccc}
\hline
\textbf{Dataset} & \textbf{ResNet-18} & \textbf{ResNet-34} & \textbf{ResNet-50} & \textbf{DenseNet-121} & \textbf{Swin-S} \\ \hline
ImageNet (LPIPS$\uparrow$) & 0.61 & 0.63 & 0.64 & 0.64 & 0.61 \\
CUB (LPIPS$\uparrow$)      & ---  & 0.61 & 0.63 & 0.62 & ---  \\ \hline
\end{tabular}
% }
\vspace{-0.6cm}
\end{table}

One of the challenges in generative prototype discovery is mode collapse, where the model produces a single, static image for a given concept. To verify that our framework generates a rich set of visual explanations, we evaluate the intra-concept diversity using the Learned Perceptual Image Patch Similarity (LPIPS) metric. For each optimized concept channel, we sample from the prompt distribution and synthesize a set of prototype images. We then compute the average pairwise LPIPS distance across all combinations of these generated variations. As reported in Tab.~\ref{tab:combined_lpips}, our method consistently achieves high LPIPS scores. Because LPIPS functions as a perceptual distance metric, these high scores confirm that the generated prototypes capture meaningful intra-concept variance rather than collapsing into deterministic outputs.

\begin{wrapfigure}{r}{0.5\textwidth}
    \vspace{-0.9cm}
    \centering
    \includegraphics[width=1\linewidth]{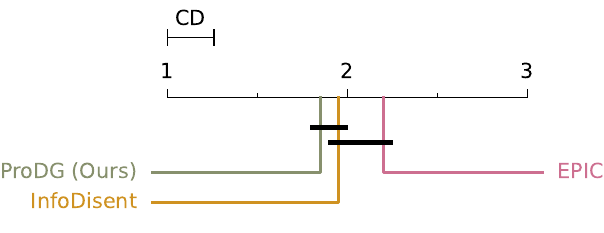}
    \caption{\textbf{Critical Difference diagram comparing user preferences for prototype-based explanation methods (\our{}, InfoDisent, and EPIC) in a three-way evaluation setting.} The diagram is based on average ranks computed from participant responses. Statistical significance is assessed using the Bonferroni-Dunn test with $\alpha = 0.05$. Methods connected by a horizontal line are not significantly different, while disconnected methods indicate a statistically significant difference.}
    \label{fig:cd_diagram}
% \end{figure}
\vspace{-0.6cm}
\end{wrapfigure}

\paragraph{User Study} We conducted a user study using an online questionnaire implemented in Google Forms to evaluate participants' perceptions and behaviors regarding the proposed method. The study included two benchmark datasets, CUB-200-2011 and ImageNet. 

The study comprised two parts. First, participants  using Likert scales evaluated individual explanations based on (i) prototype-to-input visual similarity, (ii) observability of prototype features, and (iii) visual coherence of prototype rows. Second, they compared sets of prototypes to select the one that (iv) best matched the input image, (v) best explained the recognition process across methods, and (vi) best reflected decisions across configurations.

\begin{wraptable}{r}{0.5\textwidth}
\vspace{-0.4cm}
\caption{\textbf{Users' accuracy performance in the user study on ImageNet.} The p-value is calculated for test against random selection. Due to scalability limitations, other prototype-based methods are not executable on ImageNet.}
\label{tab:user_study}
\centering
\small
\begin{tabular}{@{}l c c c@{}}
\toprule
& \textbf{\our{} (Ours)} & \textbf{EPIC} & \textbf{InfoDisent} \\
% \midrule
\cmidrule(lr){2-4}

Acc.
& $\pmb{0.73\pm0.44}$
& $0.57\pm0.5$ 
& $0.59\pm0.15$ \\[1pt]

p-value
& $3 \cdot 10^{-5}$ 
& $8 \cdot 10^{-4}$ 
& $8 \cdot 10^{-6}$ \\

\bottomrule
\end{tabular}

\vspace{-0.5cm}

\end{wraptable}

To assess prototype discriminability, participants chose between explanations initialized with strongly activated or random classes. As reported in Tab.~\ref{tab:user_study}, participants achieved significantly higher-than-chance accuracy on ImageNet. This indicates \our{} produces explanations informative enough to guide correct class identification, improving user understanding of model predictions. Extended version of results with CUB-200-2011 results are in Appendix~\ref{app:user_study}.

Participants also compared \our{}, InfoDisent, and EPIC on how well they reflected the model's decision process. Using a Critical Difference diagram (Bonferroni-Dunn, $\alpha = 0.05$), the average ranks were 1.85 for \our{}, 1.95 for InfoDisent, and 2.20 for EPIC. As shown in Fig.~\ref{fig:cd_diagram}, statistically significant differences exist between \our{} and InfoDisent, and between InfoDisent and EPIC, whereas no significant connection is found between \our{} and EPIC under this test. 

% As shown in Fig.~\ref{fig:cd_diagram}, statistically significant differences are observed between \our{} and InfoDisent, as well as between InfoDisent and EPIC. In contrast, no CD connection is found between \our{} and EPIC, indicating that their difference is not statistically significant under the chosen test. 

\textbf{Ablation Studies} To assess the contribution of each component in our objective, we perform an ablation study over all combinations of the loss terms. Specifically, our full model optimizes 
\[
\mathcal{L} = -\mathcal{L}_U + \lambda_{\text{reg}} \mathcal{L}_{\text{reg}} + \lambda_{\text{div}} \mathcal{L}_{\text{div}}.
\]
We consider all non-empty subsets of $\{ \mathcal{L}_U, \mathcal{L}_{\text{reg}}, \mathcal{L}_{\text{div}} \}$, yielding seven variants. Each variant is defined by its active loss terms: (i) $\mathcal{L}_U$, (ii) $\mathcal{L}_U + \mathcal{L}_{\text{reg}}$, (iii) $\mathcal{L}_U + \mathcal{L}_{\text{div}}$, (iv) $\mathcal{L}_{\text{reg}}$, (v) $\mathcal{L}_{\text{div}}$, (vi) $\mathcal{L}_{\text{reg}} + \mathcal{L}_{\text{div}}$, and (vii) the full objective. This setup allows us to quantify the contribution of each term to both purity and diversity (Intra-Concept and Intra-Prototype)

\begin{figure}[t]
\centering
\begin{subfigure}{0.33\linewidth}
    \includegraphics[width=\linewidth, trim=0 0 20cm 0, clip]{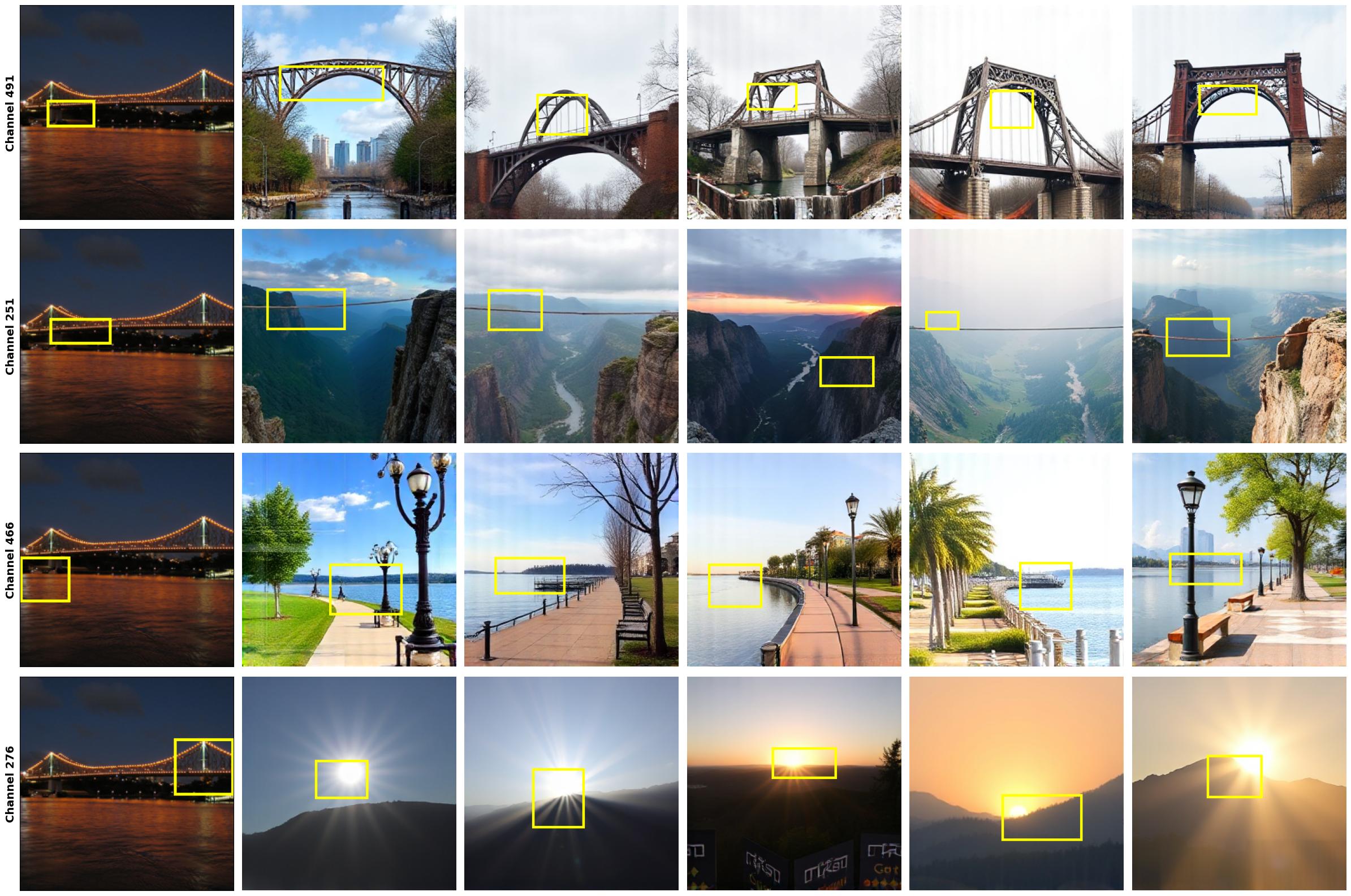}
    \caption{$\mathcal{L}_U + \mathcal{L}_{\text{reg}} + \mathcal{L}_{\text{div}}$}
    \label{fig:ablation_subfig_full}
\end{subfigure}\hfill
\begin{subfigure}{0.33\linewidth}
    \includegraphics[width=\linewidth, trim=0 0 20cm 0, clip]{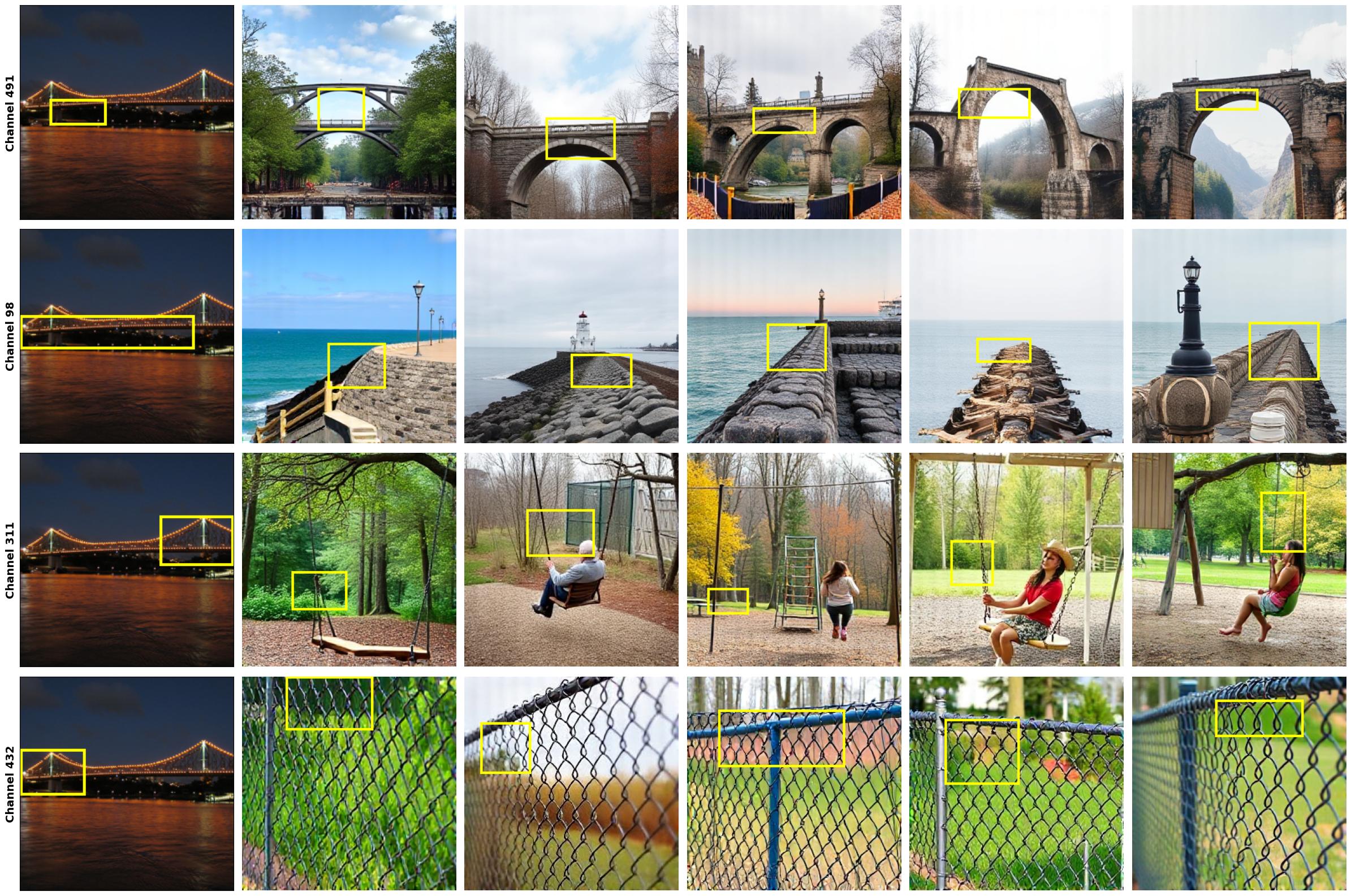}
    \caption{$\mathcal{L}_U + \mathcal{L}_{\text{div}}$}
    \label{fig:ablation_subfig_purity_div}
\end{subfigure}\hfill
\begin{subfigure}{0.33\linewidth}
    \includegraphics[width=\linewidth, trim=0 0 20cm 0, clip]{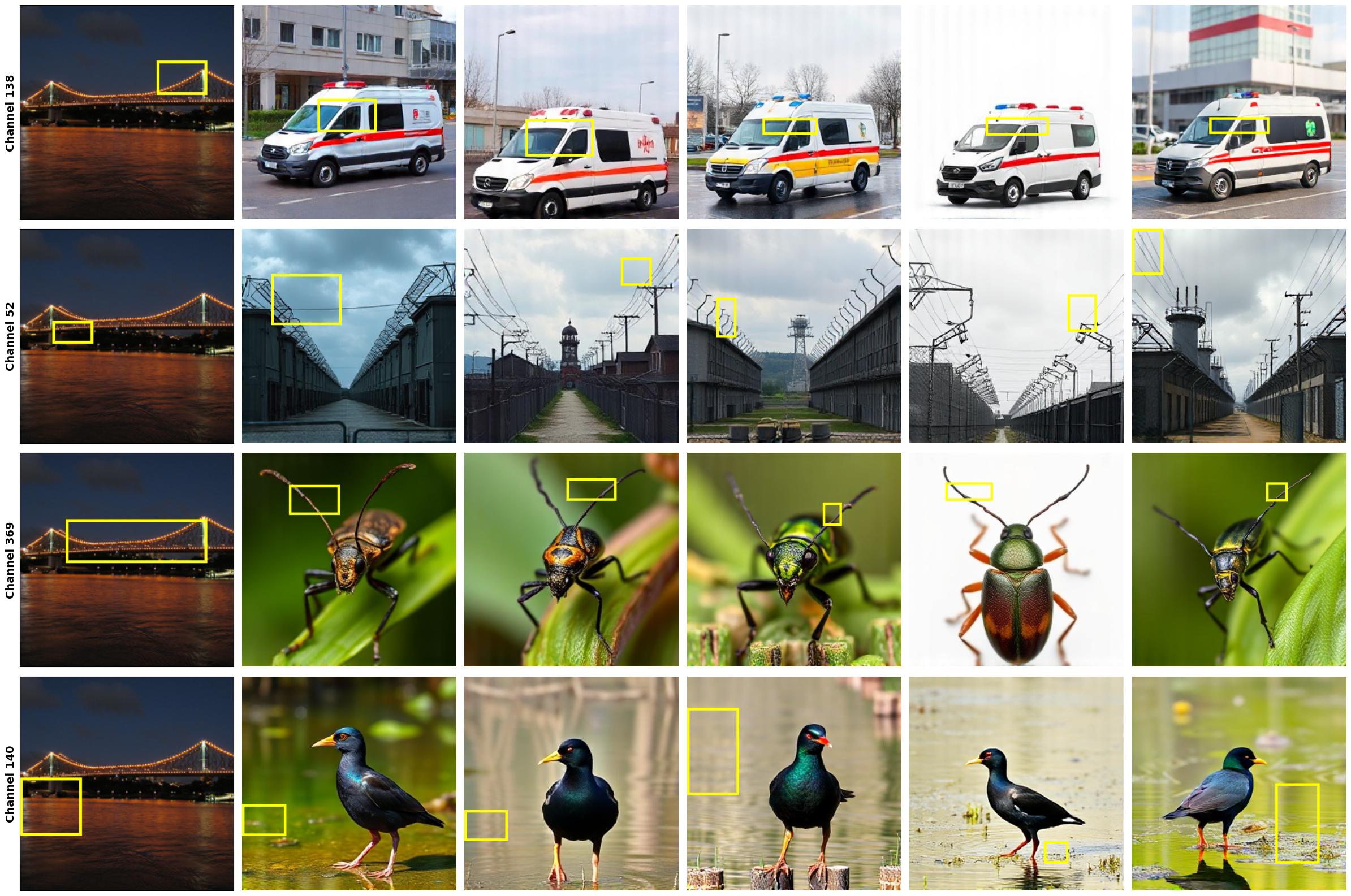}
    \caption{$\mathcal{L}_{\text{reg}} + \mathcal{L}_{\text{div}}$}
    \label{fig:ablation_subfig_reg_div}
\end{subfigure}

\caption{\textbf{Qualitative ablation study over loss components.} Each column shows samples generated under different optimization objectives. The first column shows the full model optimized with $\mathcal{L} = -\mathcal{L}_U + \lambda_{\text{reg}} \mathcal{L}_{\text{reg}} + \lambda_{\text{div}} \mathcal{L}_{\text{div}}$, while the second and third columns remove $\mathcal{L}_{\text{reg}}$ and $\mathcal{L}_U$, respectively. The full objective enforces both semantic alignment and diversity. Removing $\mathcal{L}_{\text{reg}}$ leads to weaker constraints on the prompt embeddings, resulting in degraded semantic consistency. In contrast, removing $\mathcal{L}_U$ (while adding $\mathcal{L}_{\text{div}}$) disrupts concept alignment, leading to semantic drift The full model achieves the best balance between faithful concept representation and sample diversity. }
\label{fig:ablation}
\end{figure}

Fig.~\ref{fig:ablation} presents qualitative differences across selected loss configurations. Qualitative results for all configurations are provided in the Appendix~\ref{app:ablation}. We observe that removing $\mathcal{L}_U$ leads to semantic drift, while using $\mathcal{L}_U$ combined with $\mathcal{L}_{\text{div}}$ results in poor concept alignment. The full objective achieves the best trade-off between consistency and diversity. This highlights the complementary roles of alignment ($\mathcal{L}_U$), embedding regularization ($\mathcal{L}_{\text{reg}}$), and diversity modeling ($\mathcal{L}_{\text{div}}$).

% ================================================
% ================================================
% ================================
% ====================
% ==========
% =====
% ==
% =

\section{Conclusions}

In this work, we introduced \our{}, a novel framework that enables data-free post-hoc explainability. Through a purity-optimized conditioning mechanism, we demonstrated that generative models can be successfully optimized to synthesize high-fidelity, human-interpretable visual concepts directly from the weights of a frozen classifier combined with the orthogonal feature disentanglement. This mechanism mathematically guarantees absolute faithfulness to the classifier's original predictions due to the integration of the inverse transformation within the classification head.
% Furthermore, by optimizing a robust Concept Prompt Bank rather than static image patches, our approach naturally captures intra-concept visual diversity and prevents mode collapse, allowing for precise spatial localization of concepts on novel inputs.
By providing a secure, privacy-preserving alternative for model interpretation, \our{} enables transparency whenever access to external data is restricted.

\textbf{Limitations and Future Work}
While our data-free approach resolves the critical issue of data dependency, it introduces constraints inherited from generative models. Specifically, the framework is affected by generative prior bias, if the target concepts are entirely out-of-distribution for the underlying model, the prompt optimization may struggle to synthesize valid visual prototypes. Consequently, applying the framework to specialized domains requires swapping to a domain-specific generative model. Additionally, while our text anchor initialization improves convergence, more advanced prompt initializations could further enhance the capture of highly abstract visual features.

% \textbf{Limitations and Future Work}
% While our data-free approach resolves the critical issue of data dependency, it introduces constraints inherited from generative models. Consequently, applying the framework to specialized domains simply requires swapping to a domain-specific generative model. Additionally, while our text anchor initialization improves convergence, more advanced prompt initializations could further enhance the capture of highly abstract visual features.

% \begin{ack}
% Use unnumbered first level headings for the acknowledgments. All acknowledgments

% Do {\bf not} include this section in the anonymized submission, only in the final paper. You can use the \texttt{ack} environment provided in the style file to automatically hide this section in the anonymized submission.
% \end{ack}

\medskip

\bibliographystyle{unsrt}
%\bibliography{ref}

% {
% \small
% [1] Alexander, J.A.\ \& Mozer, M.C.\ (1995) Template-based algorithms for
% connectionist rule extraction. In G.\ Tesauro, D.S.\ Touretzky and T.K.\ Leen
% (eds.), {\it Advances in Neural Information Processing Systems 7},
% pp.\ 609--616. Cambridge, MA: MIT Press.

% [2] Bower, J.M.\ \& Beeman, D.\ (1995) {\it The Book of GENESIS: Exploring
%   Realistic Neural Models with the GEneral NEural SImulation System.}  New York:
% TELOS/Springer--Verlag.

% [3] Hasselmo, M.E., Schnell, E.\ \& Barkai, E.\ (1995) Dynamics of learning and
% recall at excitatory recurrent synapses and cholinergic modulation in rat
% hippocampal region CA3. {\it Journal of Neuroscience} {\bf 15}(7):5249-5262.
% }

%%%%%%%%%%%%%%%%%%%%%%%%%%%%%%%%%%%%%%%%%%%%%%%%%%%%%%%%%%%%
% ================================================
% ================================================
% ================================
% ====================
% ==========
% =====
% ==
% =

\clearpage

\appendix

\section{User Study}
\label{app:user_study}
\begin{figure}
    \centering
    \includegraphics[width=\linewidth]{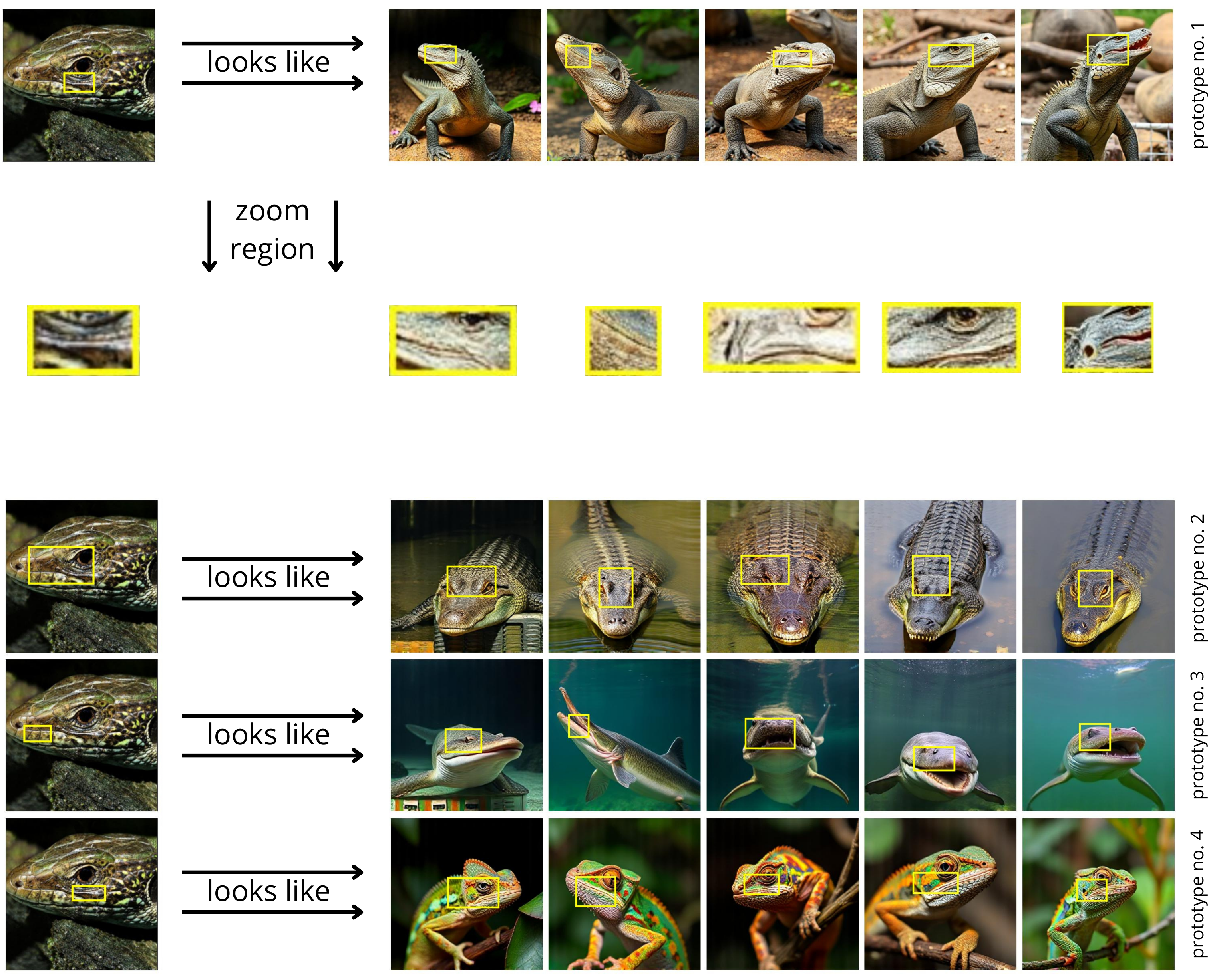}
    \caption{\textbf{User study instructions and guide.} Illustrative guide presented prior to the user-study questionnaire, demonstrating the concept of a prototype and how to interpret prototype-based explanations. The figure provides an intuitive guide for understanding visualizations in our framework.}
    \label{fig:user_study_exaple}
\end{figure}

\begin{table}[t]
\caption{\textbf{Performance in task (v) of the user study.} The table reports accuracy performance, standard deviation, and p-values for in the user study. The p-value column indicates the significance of a test against random selection.}
\label{tab:user_study_app}
\centering
\small
{
\setlength{\tabcolsep}{3.3pt}
\footnotesize
\begin{NiceTabular}{@{}ll c c c c c c c@{}}
\toprule
\textbf{Dataset} &  & \textbf{\our{} (Ours)} & \textbf{EPIC} & \textbf{InfoDisent} & \textbf{ProtoPNet} & \textbf{ProtoConcepts} & \textbf{PIPNet} & \textbf{LucidPPN} \\
\midrule
% \cmidrule(lr){3-9}

\Block{2-1}{ImageNet}
% \multirow{2}{*}{\textit{\rotatebox{75}{ImageNet}}}
% & \textbf{\rotatebox{75}{Acc}} 
& Acc
& $0.73\pm0.44$ 
& $0.57\pm0.5$ 
& $0.59\pm0.15$ 
& -- 
& -- 
& -- 
& -- \\[1pt]

% & \textbf{\rotatebox{75}{p-val}} 
& p-val
& $3 \cdot 10^{-5}$ 
& $8 \cdot 10^{-4}$ 
& $8 \cdot 10^{-6}$ 
& -- 
& -- 
& -- 
& -- \\

\midrule
\Block{2-1}{CUB}
% \multirow{2}{*}{\textit{\rotatebox{75}{CUB}}}
% & \textbf{\rotatebox{75}{Acc}} 
& Acc
& $0.37\pm0.48$ 
& $0.55\pm0.5$ 
& $0.65\pm0.13$ 
& $0.52\pm0.05$ 
& $0.62\pm0.05$ 
& $0.60\pm0.18$ 
& $0.68\pm0.17$ \\[1pt]

% & \textbf{\rotatebox{75}{p-val}} 
& p-val
& $0.99$ 
& $9 \cdot 10^{-3}$ 
& $10^{-14}$ 
& $0.29$ 
& $3 \cdot 10^{-5}$ 
& $0.002$ 
& $2 \cdot 10^{-6}$ \\

\bottomrule
\end{NiceTabular}
}
\end{table}

In our user study, each question included multiple examples to ensure robustness, with participants evaluating three instances from ImageNet and three from CUB-200-2011. The study was conducted anonymously, without financial compensation and without time constraints. Prior to the main task, participants were provided with detailed instructions and illustrative guide, as presented in Fig.~\ref{fig:user_study_exaple}, to familiarize them with the explanation format and evaluation criteria. 

\begin{figure}[t]
    \centering
    \begin{NiceTabular}{X[c]@{\extracolsep{\fill}}X[c]}
    \Block[]{2-1}{}
    \includegraphics[width=0.9\textwidth, trim=2.7cm 5.5cm 3cm 3cm, clip]{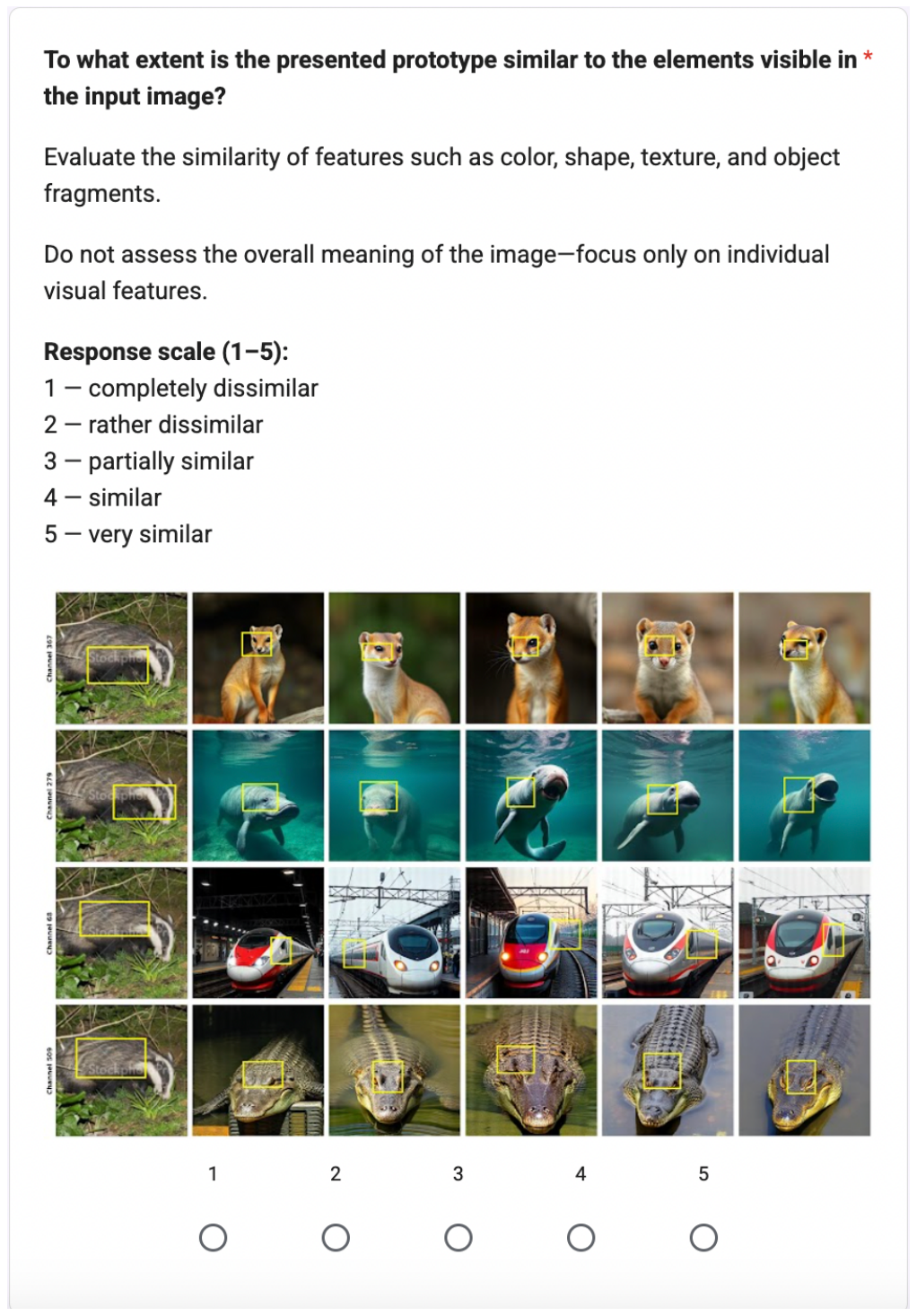}
    &
    \Block[]{2-1}{}
    \includegraphics[width=0.9\textwidth, trim=2.7cm 6.5cm 3cm 3cm, clip]{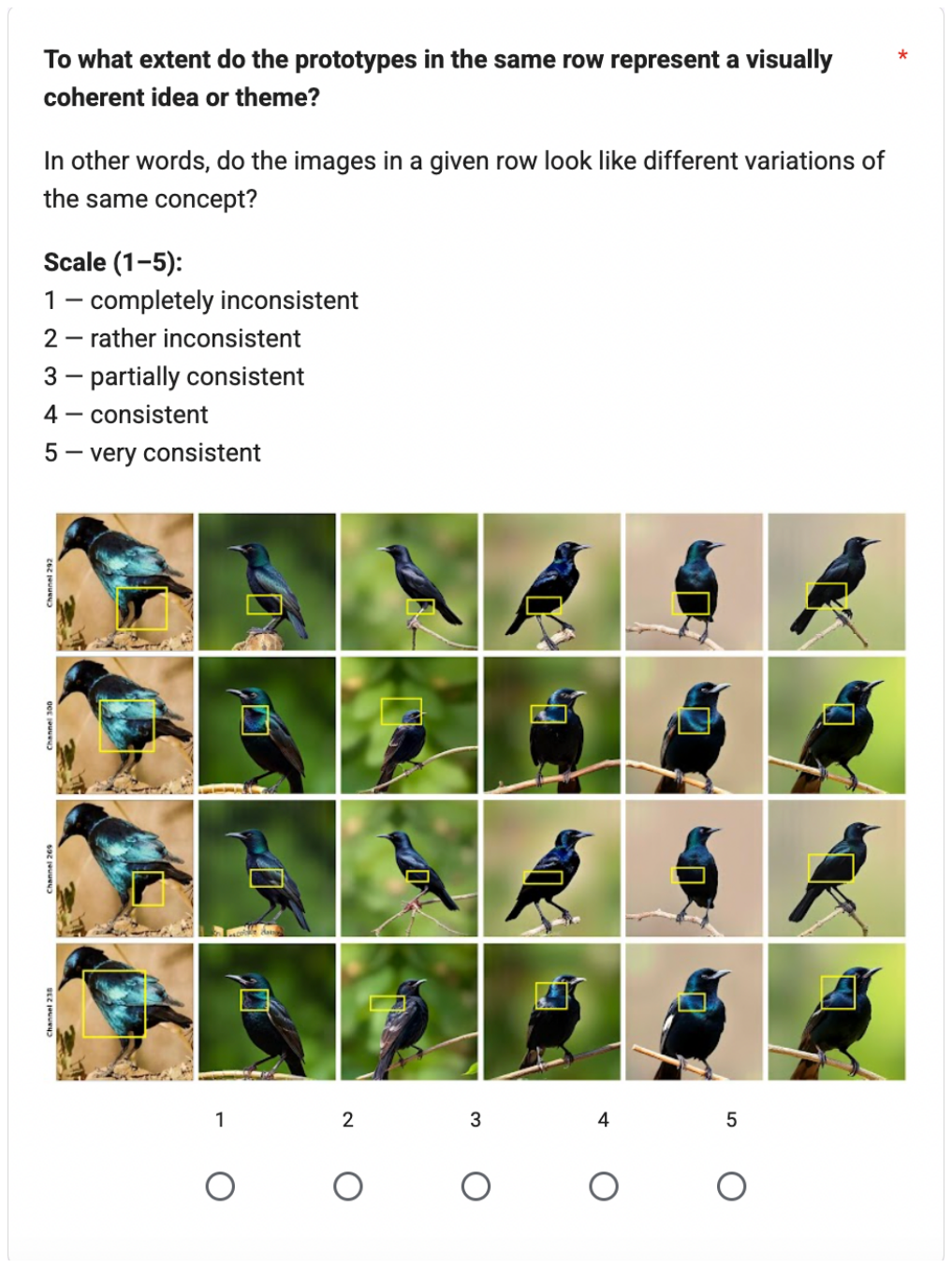}
    \\
    (i) & (ii)
    \end{NiceTabular} 
    
    \caption{\textbf{Example questions (i) and (ii) from the first part of the user study.} Participants were asked to evaluate the visual similarity between the generated prototype and the input image, as well as the visual coherence of the prototypes within a row, using a 1-5 Likert scale.}
    \label{fig:user_study_part1_ex2}
\end{figure}

\begin{figure}[t]
    \centering
    \includegraphics[width=0.55\linewidth, trim=2.7cm 6.5cm 3cm 3cm, clip]{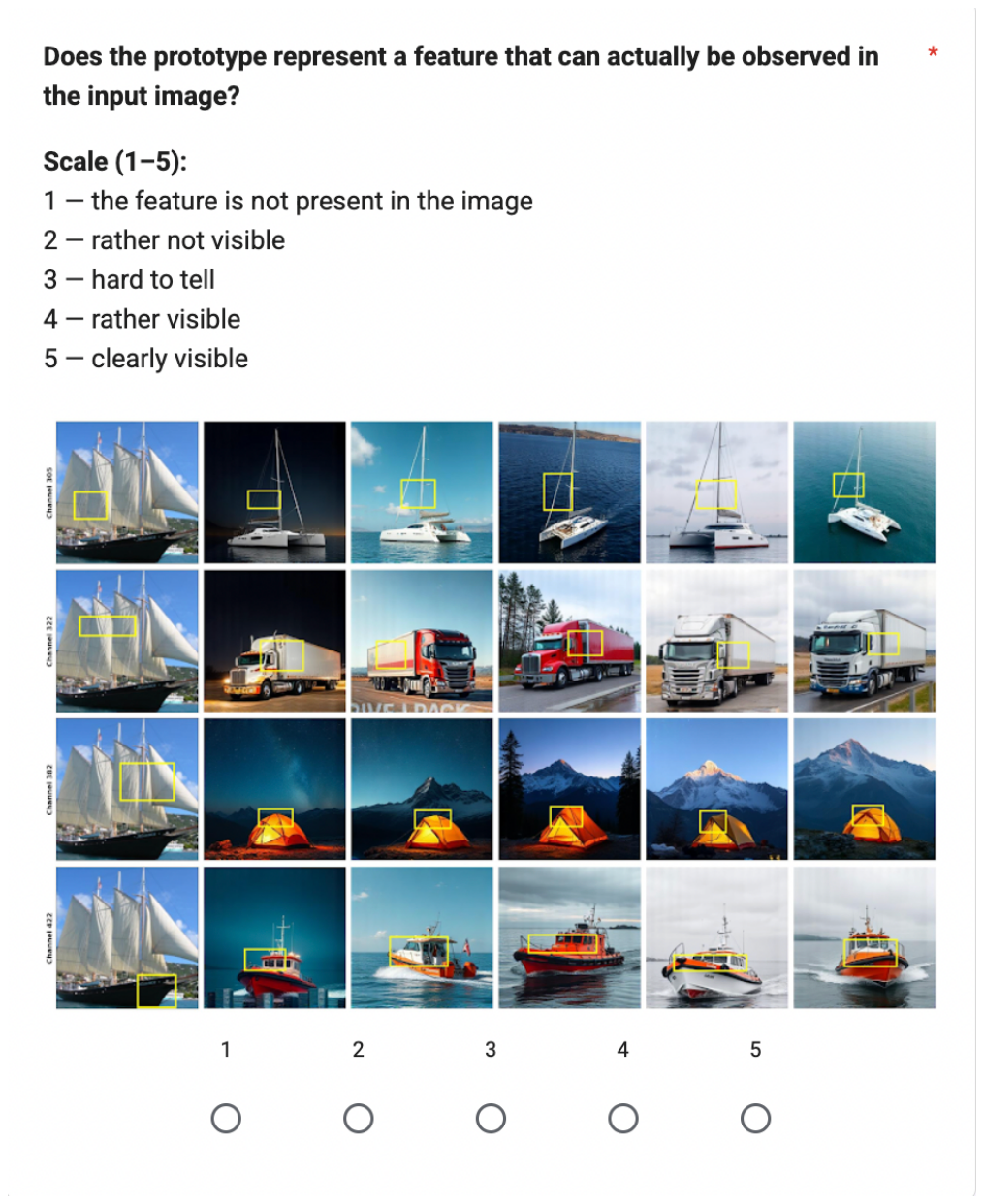}
    \caption{\textbf{Example question (iii) from the first part of the user study.} Participants evaluated whether the specific concepts highlighted by the generative prototypes can actually be observed in the original input image.}
    \label{fig:user_study_part1_ex3}
\end{figure}

The fourth question in the user study aimed to assess how well participants could differentiate between prototypical parts and rely on them for downstream classification. Participants were shown two types of explanations: one initialized using the most strongly activated class names and another based on randomly selected classes. Their task was to select which explanation better matched the input image, relying only on the provided explanations. 

As reported in Tab.~\ref{tab:user_study_app}, participants achieved significantly higher-than-chance accuracy on ImageNet. For the CUB-200-2011 dataset, we observe lower accuracy, which is expected given the fine-grained nature of the classes and their strong visual similarity. Despite this increased difficulty, the results still suggest that the explanations retain useful discriminative information.

\begin{figure}[t]
    \centering
    \includegraphics[width=0.9\linewidth, trim=1cm 8cm 1cm 8cm, clip]{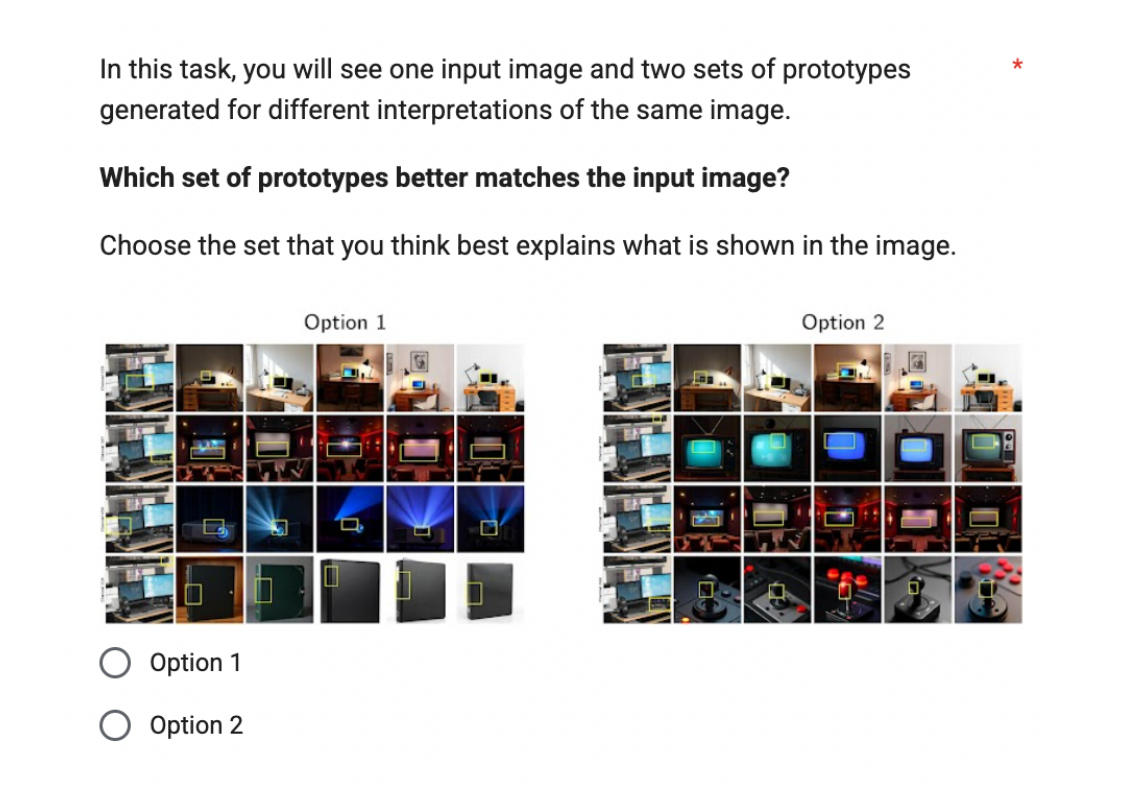}
    \caption{\textbf{Example question (iv) from the second part of the user study.} Participants were presented with alternative sets of prototypes most influential for selecting different classes, including the correct one, and were asked to select the one that best explains the prediction. }
    \label{fig:user_study_part2_ex1}
\end{figure}

\begin{figure}[t]
    \centering
    \includegraphics[width=0.9\linewidth, trim=1cm 8cm 1cm 8cm, clip]{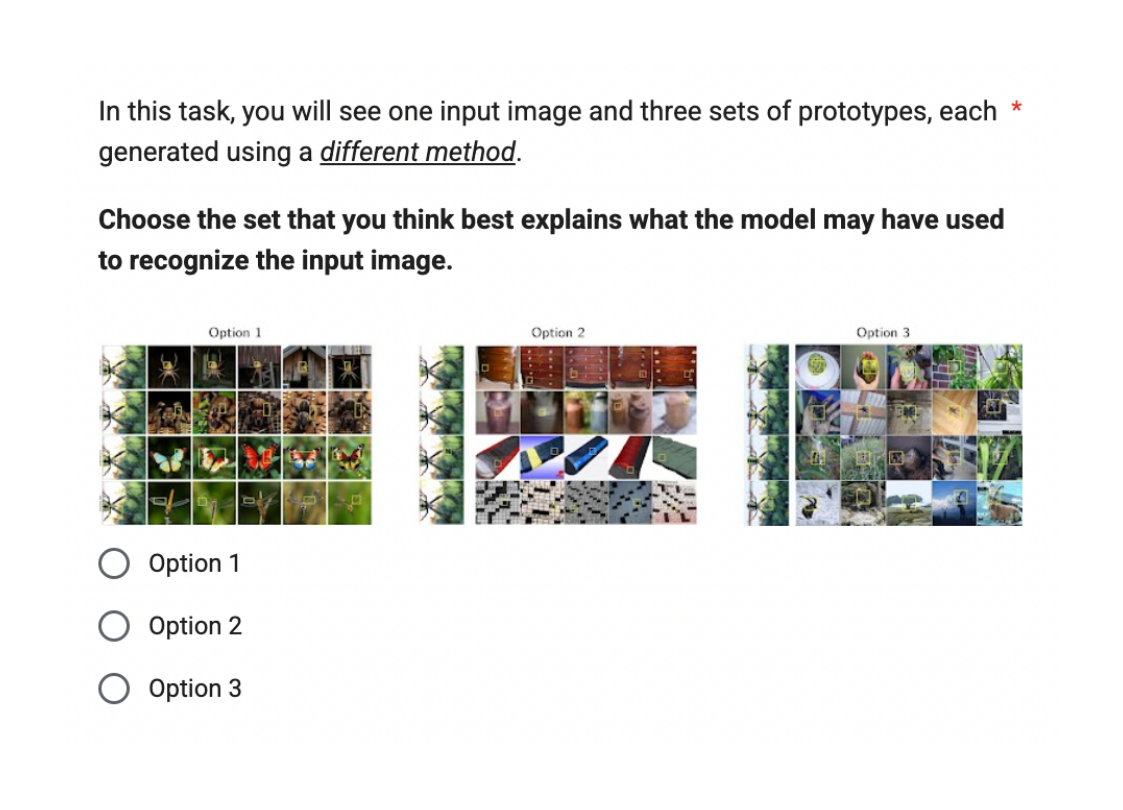}
    \caption{\textbf{Example question (v) from the second part of the user study.} Participants were presented with alternative sets of prototypes found by different models (\our{}, EPIC and InfoDisent). They were asked select the one that best explains the given input image.}
    \label{fig:user_study_part2_ex2}
\end{figure}

\begin{figure}[t]
    \centering
    \includegraphics[width=0.9\linewidth, trim=1cm 8cm 2.3cm 8cm, clip]{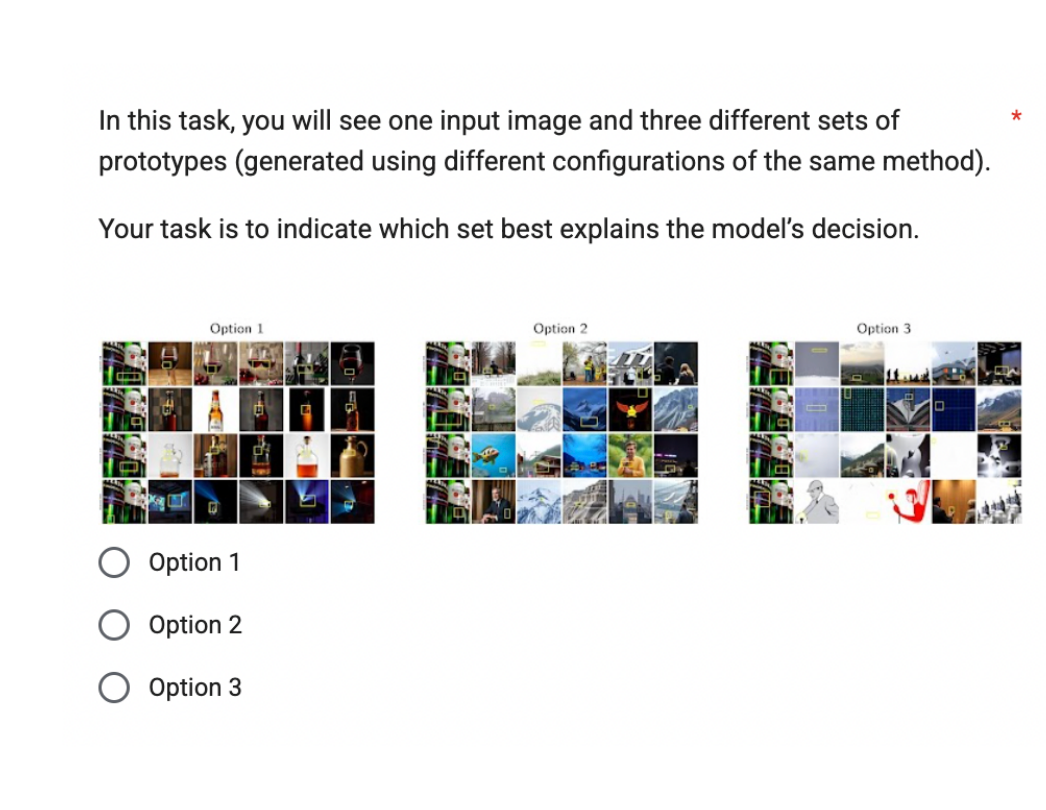}
    \caption{\textbf{Example question (vi) from the second part of the user study.} Participants chose which initialization of our method produces prototypes that best capture the defining features of the input image.}
    \label{fig:user_study_part2_ex3}
\end{figure}

To provide further context on the study's design, we present sample questions extracted directly from the user study interface. Figs~\ref{fig:user_study_part1_ex2} and~\ref{fig:user_study_part1_ex3} demonstrate the 1-5 Likert scale evaluation tasks from the first part of the user study. Figs~\ref{fig:user_study_part2_ex1},~\ref{fig:user_study_part2_ex2} and~\ref{fig:user_study_part2_ex3} illustrate the comparative multiple-choice tasks from the second part. 

\section{Explanations of model decision}
\label{app:comparisons}
\begin{figure*}[t]
    \centering
    \includegraphics[width=0.45\linewidth]{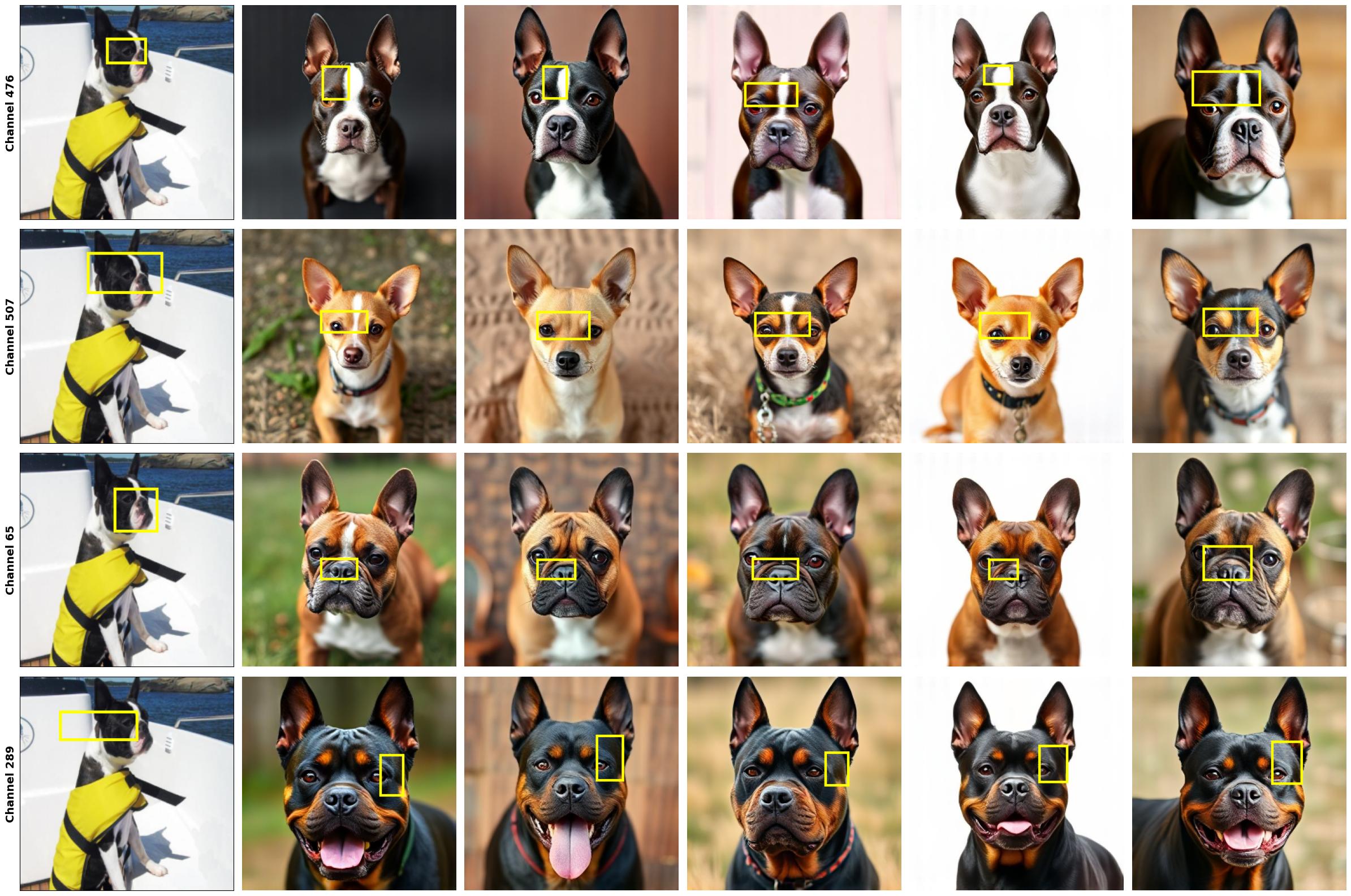}
        \quad 
        \includegraphics[width=0.45\linewidth]{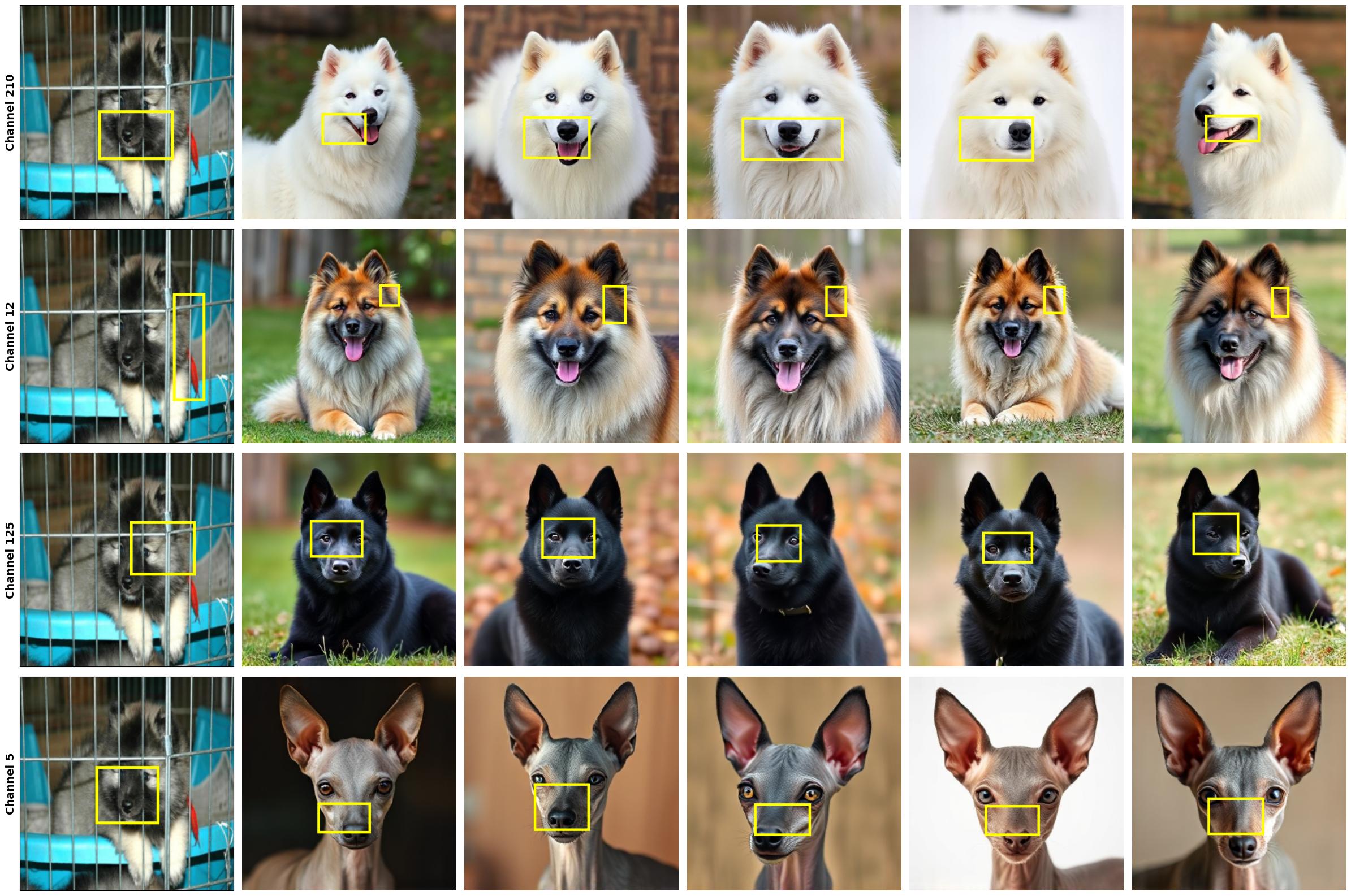} \\
    \vspace{1cm}
    \includegraphics[width=0.45\linewidth]{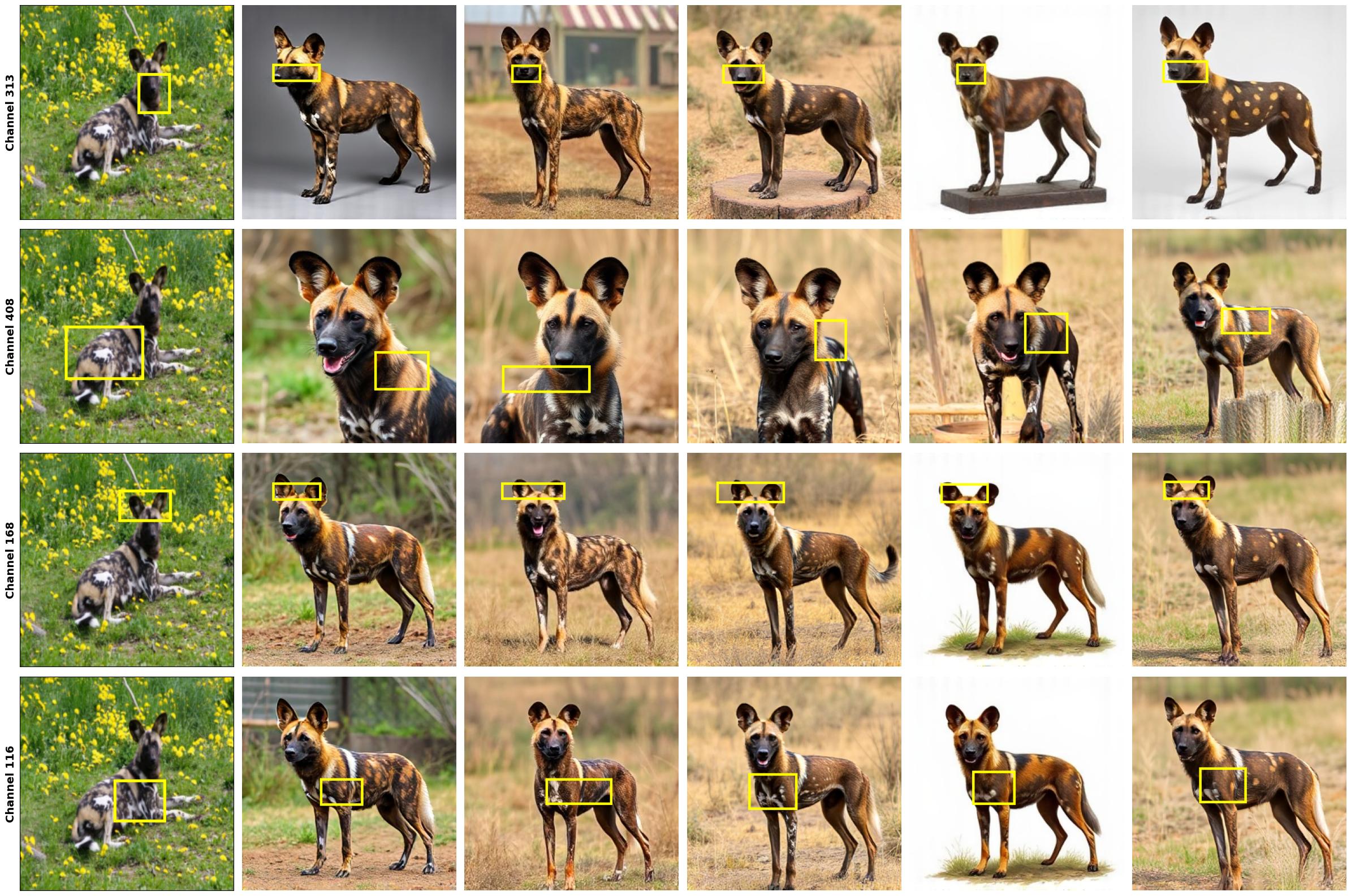}
        \quad 
        \includegraphics[width=0.45\linewidth]{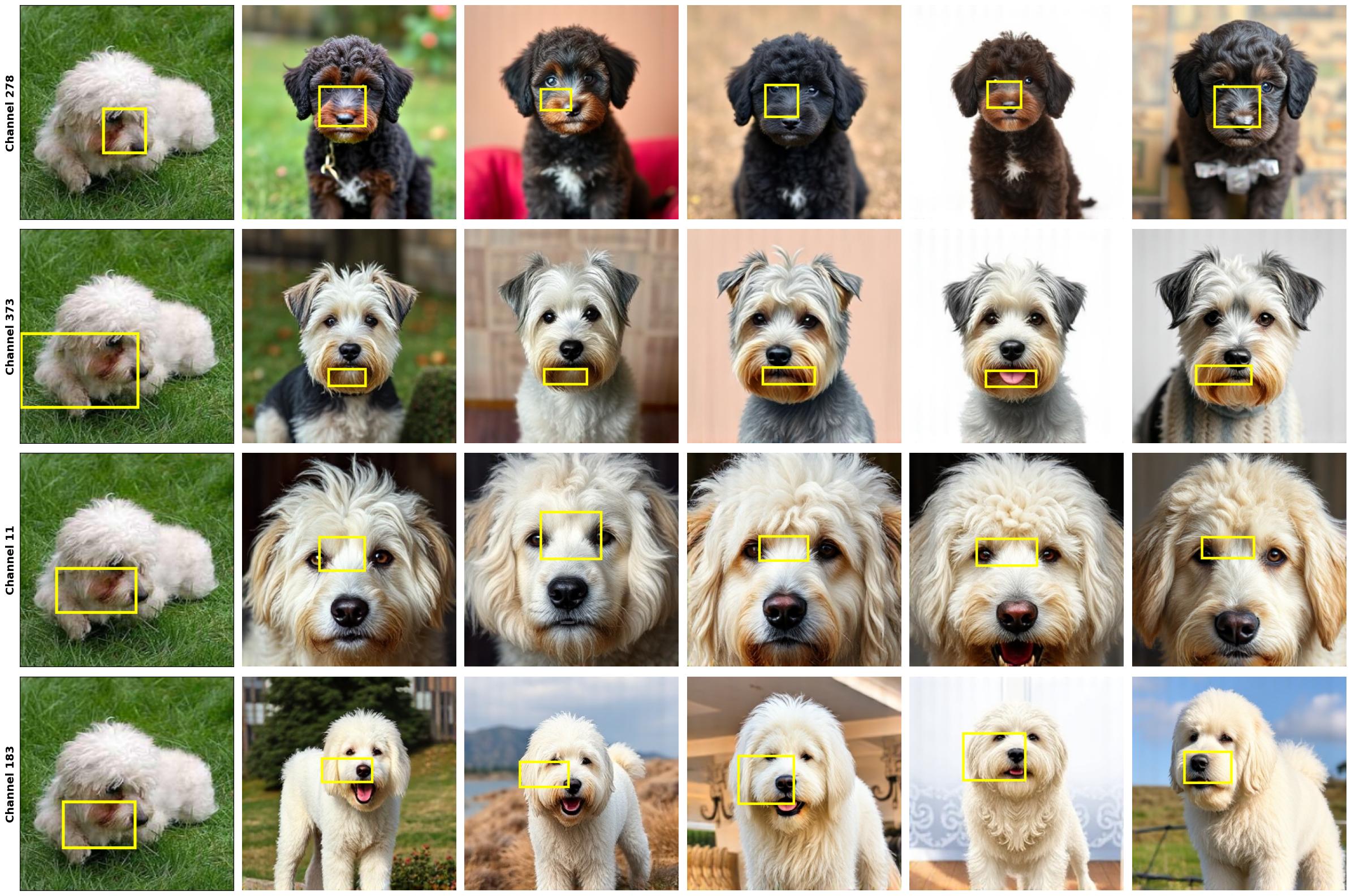} \\
    
    \caption{\textbf{Example \our{} prototypes generated for the Stanford Dogs dataset.} Each row highlights prototypes from a specific channel, focusing on different dog features such as ears, nose, and fur. Observe that the dogs' breeds observed in prototypes are similar.}
    \label{fig:stanford_dogs_examples}
\end{figure*}

\begin{figure*}[t]
    \centering
    \includegraphics[width=0.45\linewidth]{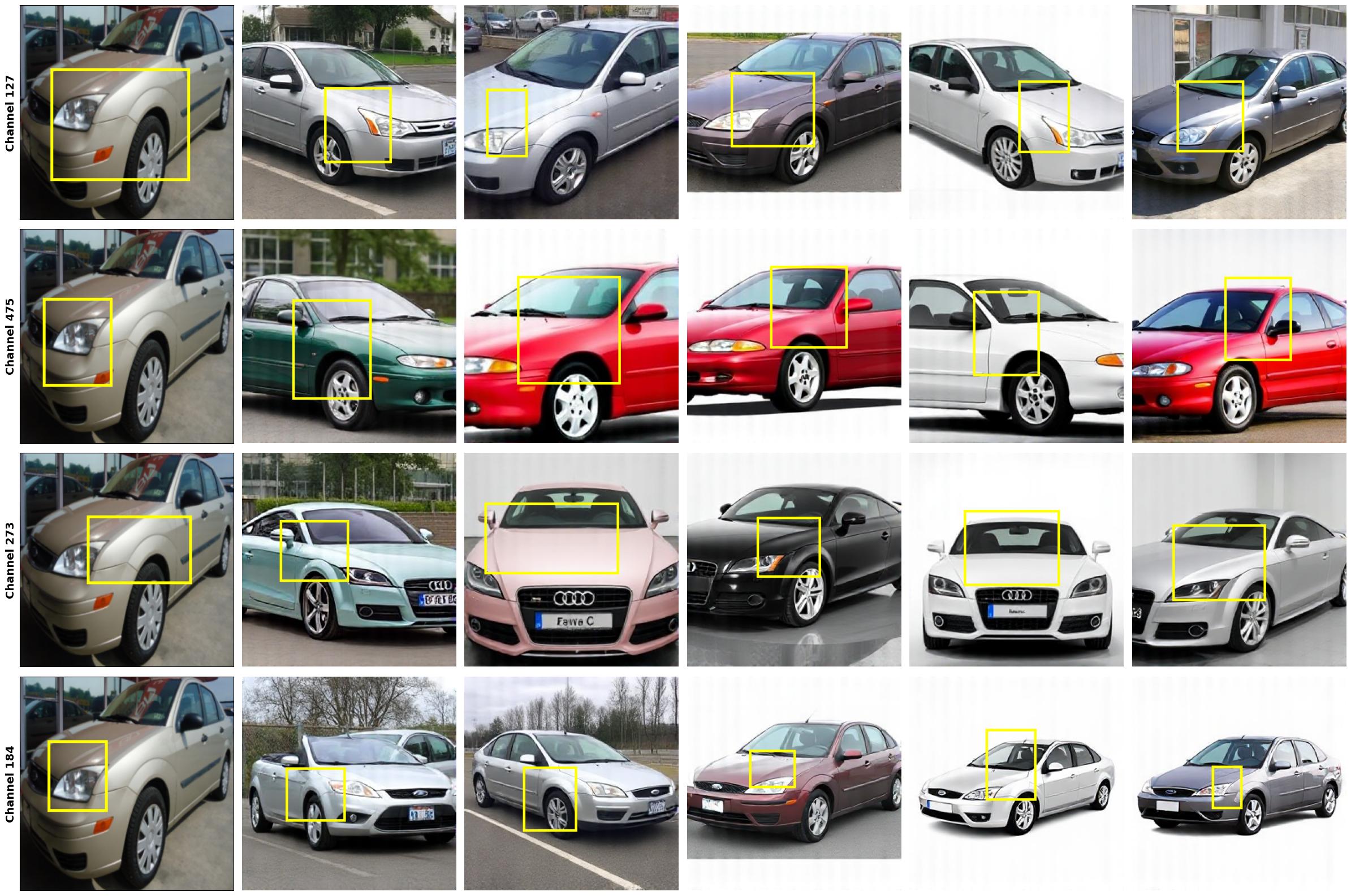}
        \quad 
        \includegraphics[width=0.45\linewidth]{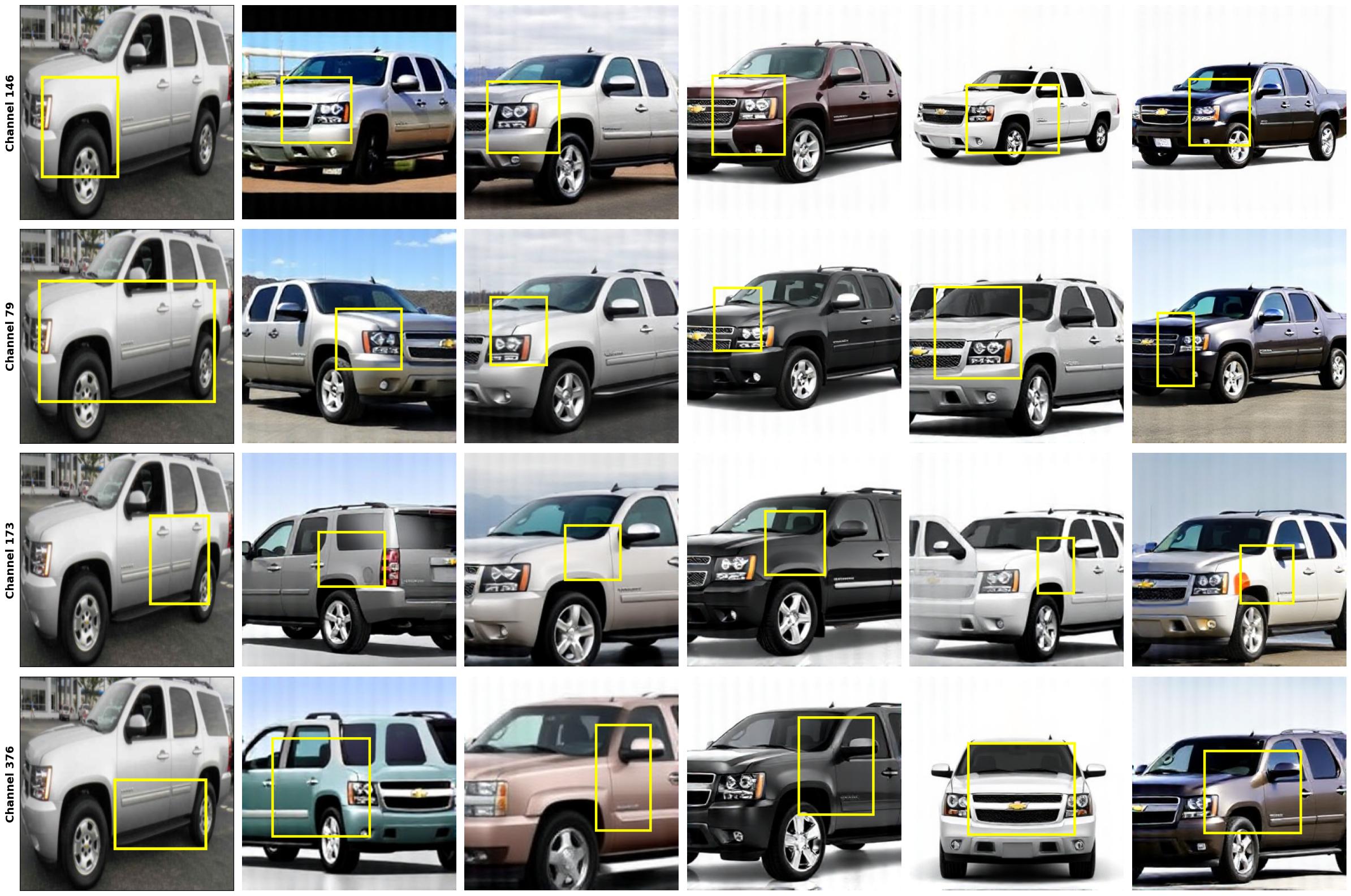} \\
    \vspace{1cm}
    \includegraphics[width=0.45\linewidth]{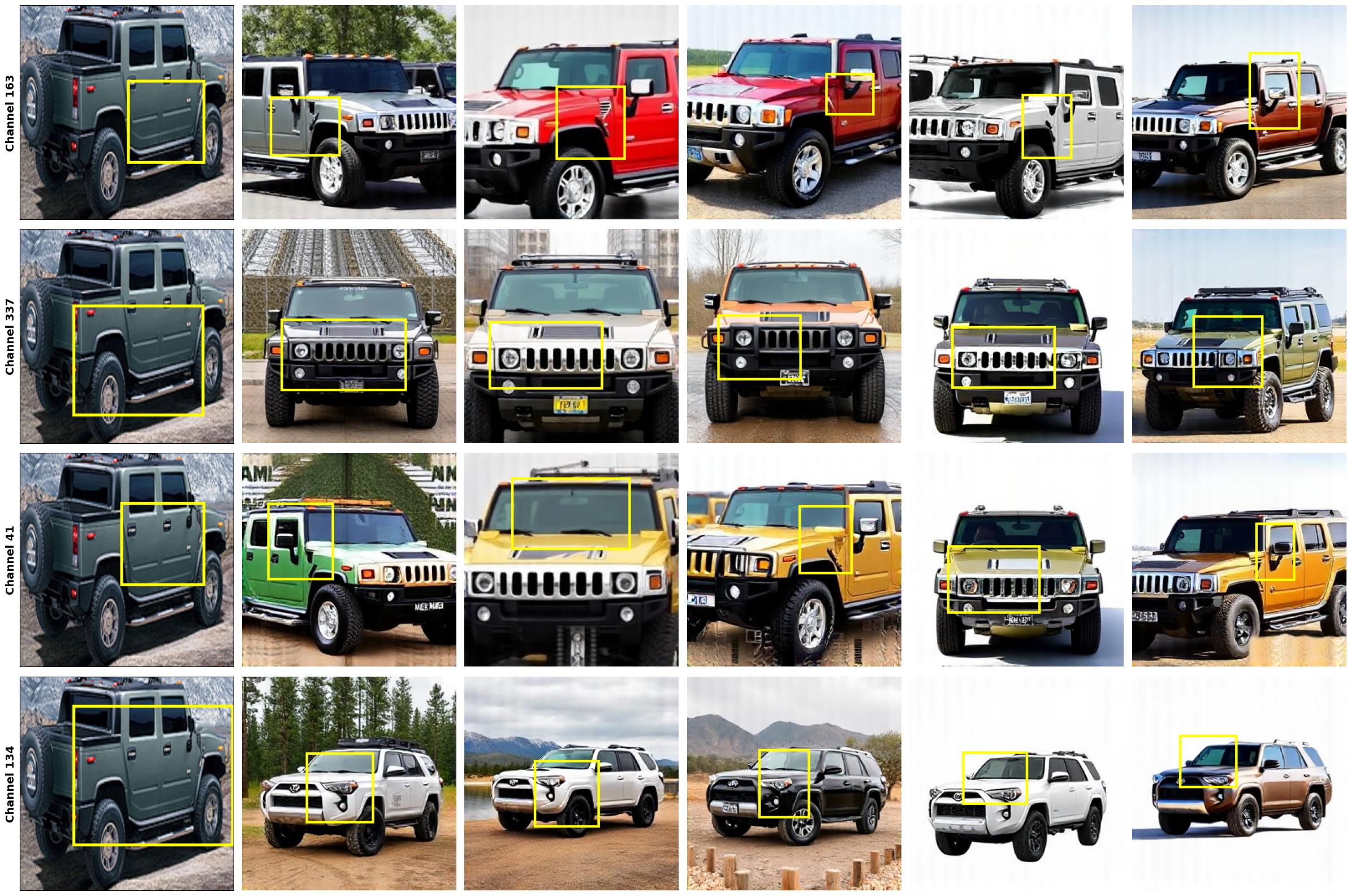}
        \quad 
        \includegraphics[width=0.45\linewidth]{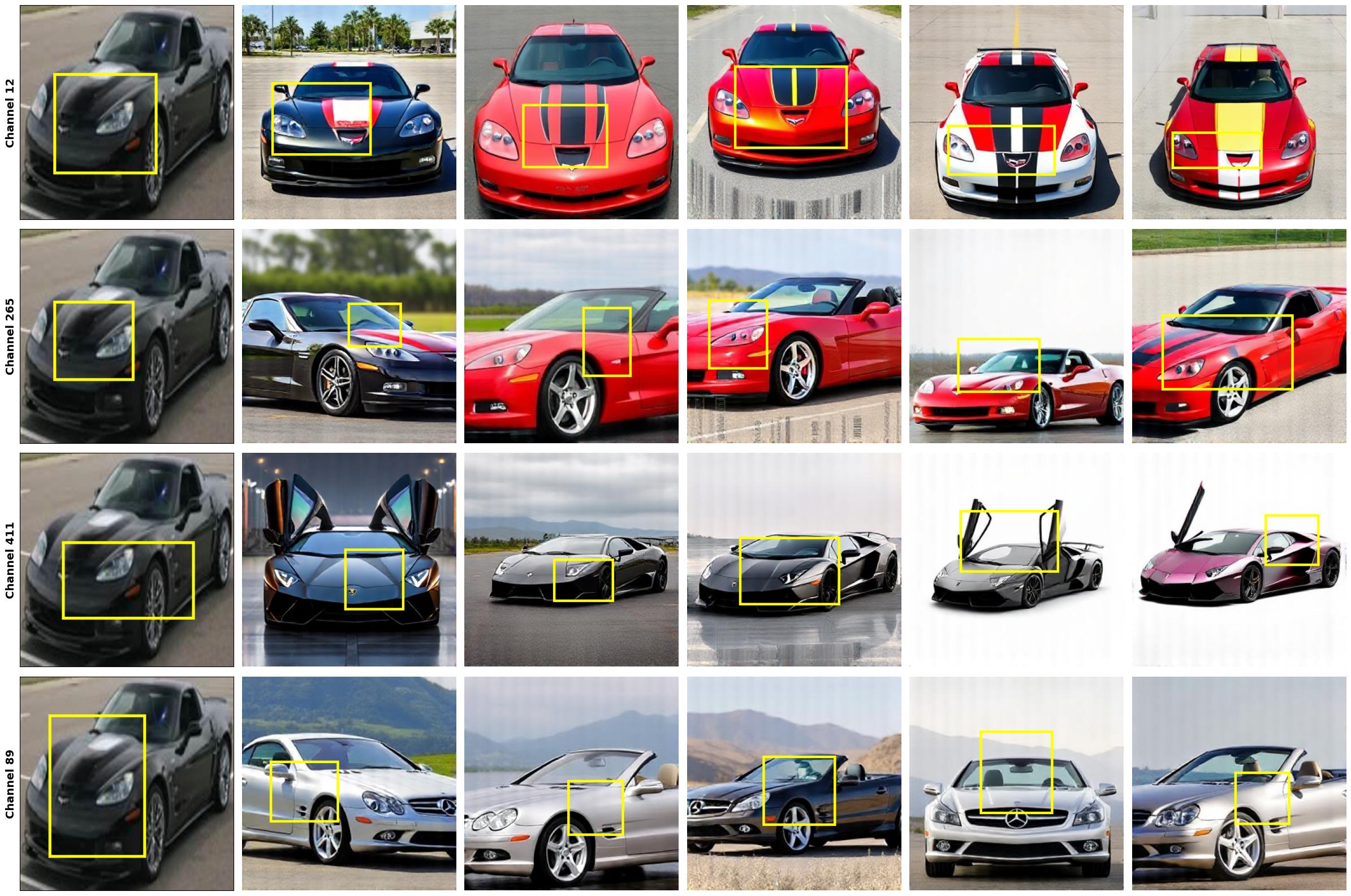} \\
    
    \caption{\textbf{Example \our{} prototypes generated for the Stanford Cars dataset.} Each row highlights prototypes from a specific channel, focusing on different part of vehicles.}
    \label{fig:stanford_cars_examples}
\end{figure*}

In this section, we provide additional results of experiments in explanations of model decision made by \our{}. To demonstrate the versatility and robustness of our generative approach across diverse visual domains, we present additional results on the fine-grained Stanford Dogs (see Fig.~\ref{fig:stanford_dogs_examples}) and Stanford Cars (see Fig.~\ref{fig:stanford_cars_examples}) datasets.

\begin{figure*}[t]
    \centering
    \our{} (Ours) \hspace{9em} EPIC \hspace{11em}  InfoDisent \\
    \includegraphics[width=\linewidth, trim= 1cm 0cm 0cm 6cm, clip]{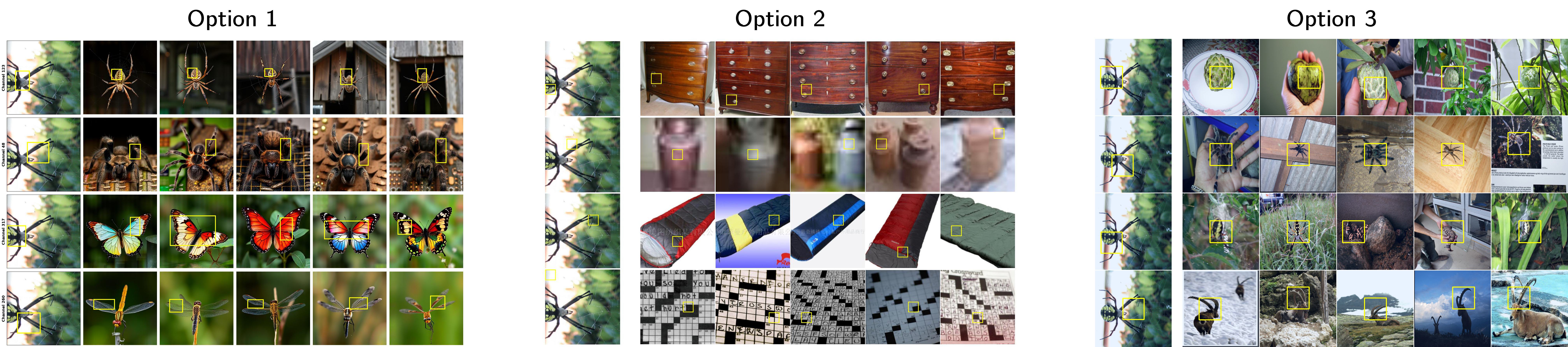}\\
    \hspace{5cm}
    \includegraphics[width=\linewidth, trim= 1cm 0cm 0cm 6cm, clip]{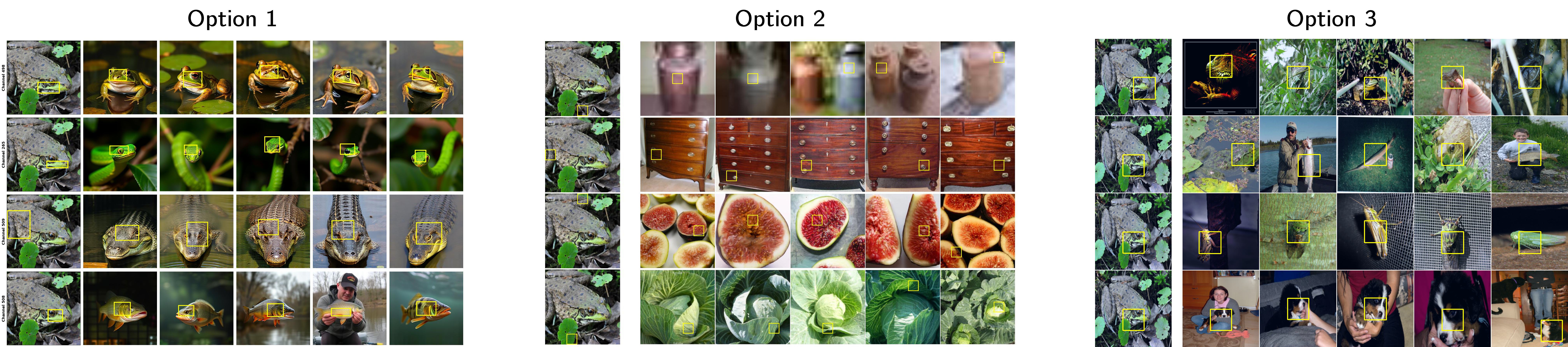}\\
    \hspace{5cm}
    \includegraphics[width=\linewidth, trim= 1cm 0cm 0cm 6cm, clip]{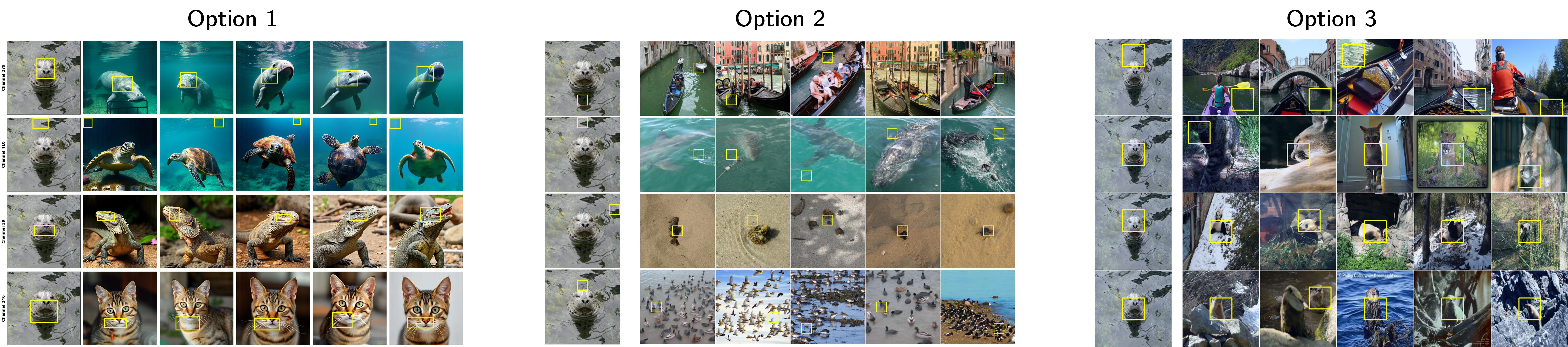}\\
    \hspace{5cm}
    \includegraphics[width=\linewidth, trim= 1cm 0cm 0cm 6cm, clip]{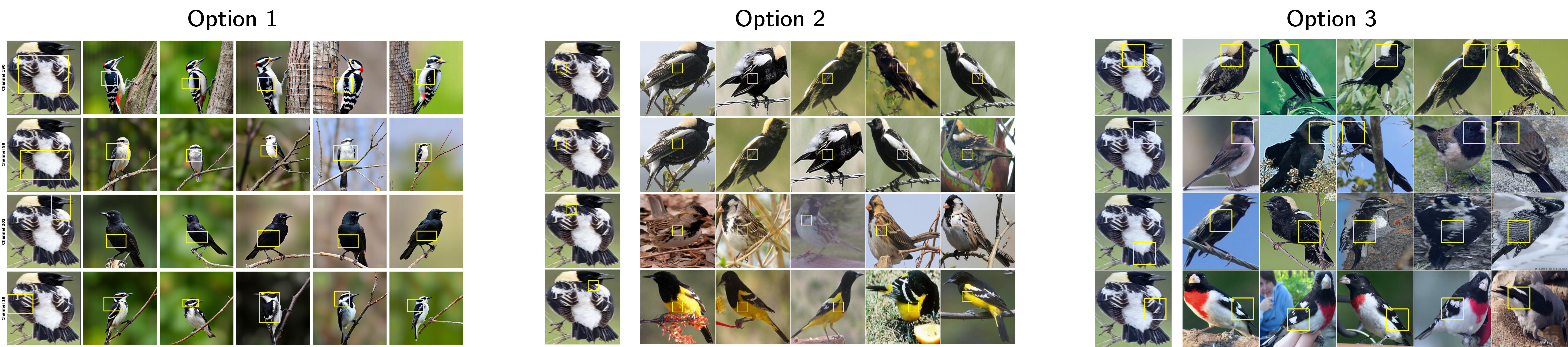}\\
    \hspace{5cm}
    \includegraphics[width=\linewidth, trim= 1cm 0cm 0cm 6cm, clip]{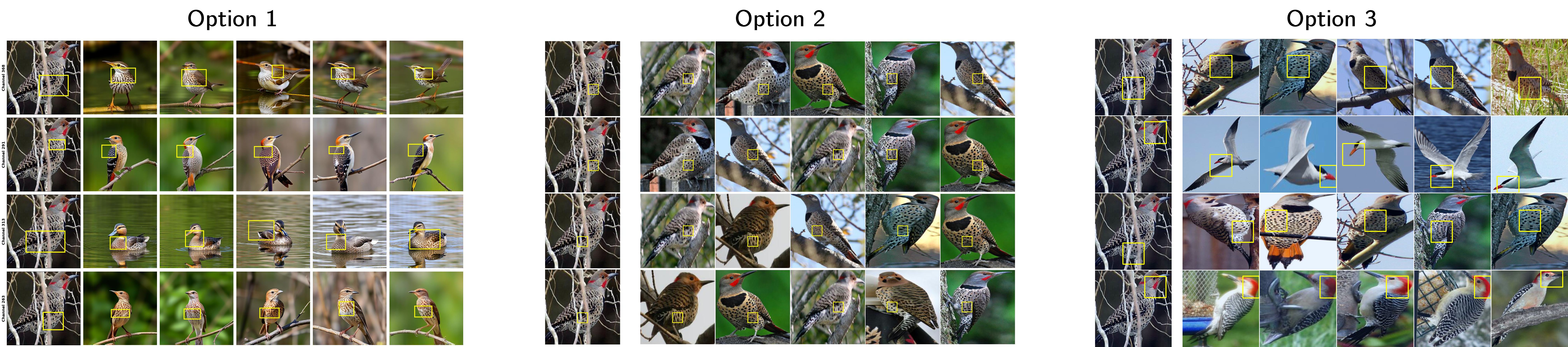}\\
    \hspace{5cm}
    \includegraphics[width=\linewidth, trim= 1cm 0cm 0cm 6cm, clip]{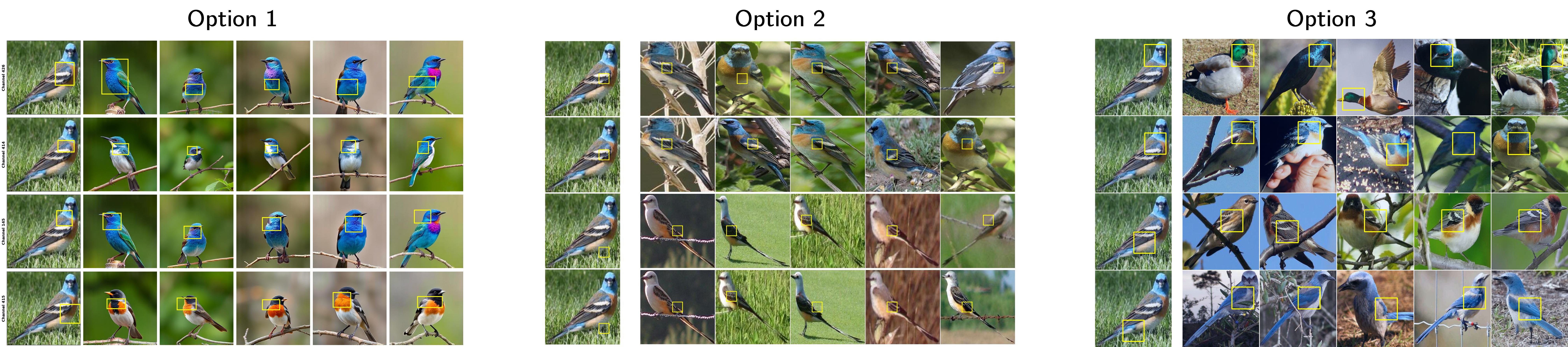}\\
    
    \caption{\textbf{Qualitative comparison of visual explanations generated by \our{} (Ours), EPIC, and InfoDisent across both CUB-200-2011 and ImageNet datasets.} The baselines utilize localized image crops to highlight features, whereas \our{} synthesizes complete images to encapsulate the learned concepts}
    \label{fig:appendix_comparison_epic_infodisent_freegen}
\end{figure*}

Furthermore, we provide additional results of explanations made by \our{} with a direct visual comparison to prototype-based mathods: EPIC and InfoDisent. Examples shown in Fig.~\ref{fig:appendix_comparison_epic_infodisent_freegen} are selected from both the ImageNet and CUB-200-2011 datasets.

\section{More details about ablation studies}
\label{app:ablation}

\begin{figure}[t]
\centering

\begin{subfigure}{0.3\linewidth}
    \includegraphics[width=\linewidth, trim=0 0 20cm 0, clip]{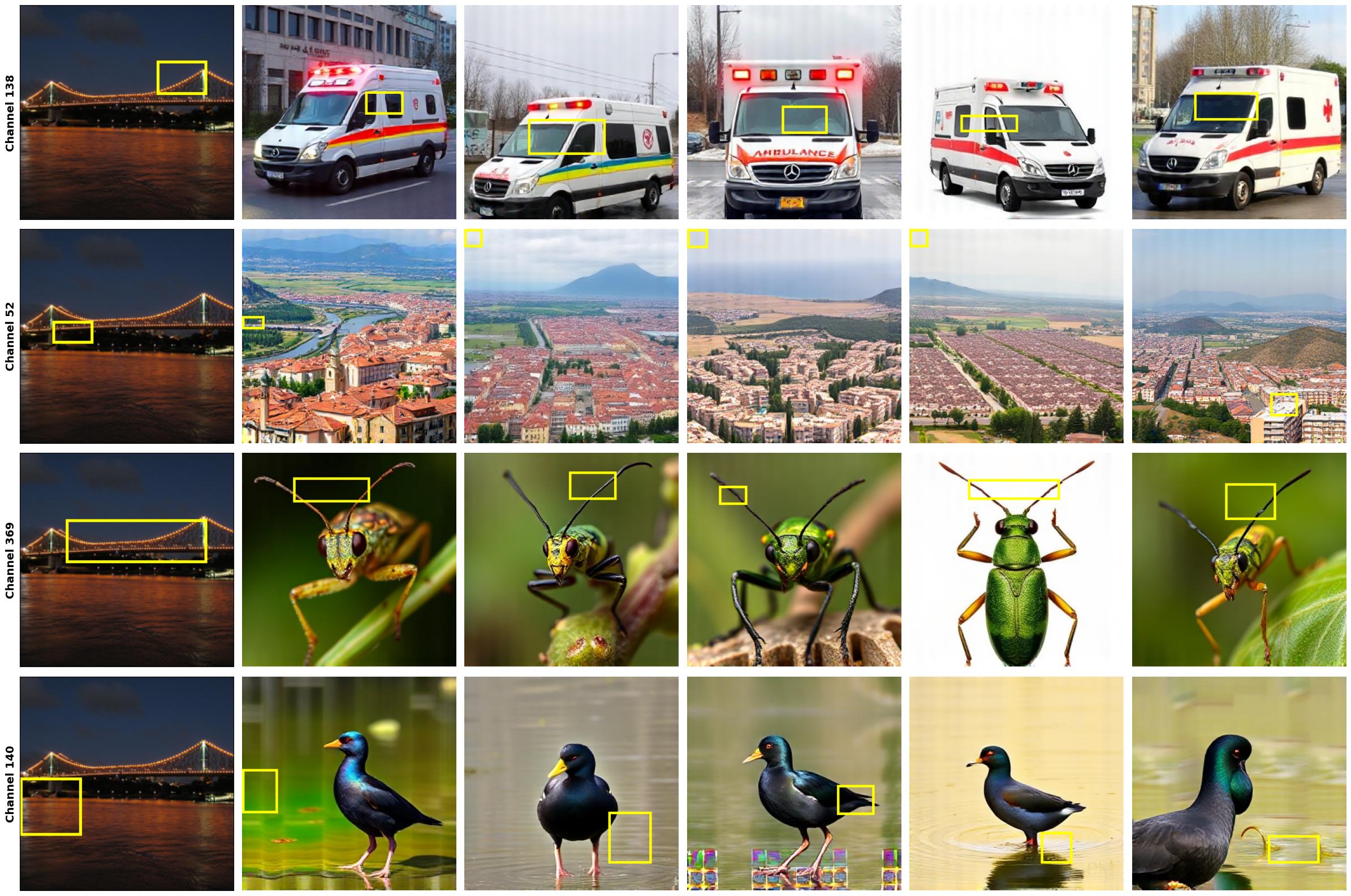}
    \caption{$\mathcal{L}_{\text{div}}$}
\end{subfigure}\hfill
\begin{subfigure}{0.3\linewidth}
    \includegraphics[width=\linewidth, trim=0 0 20cm 0, clip]{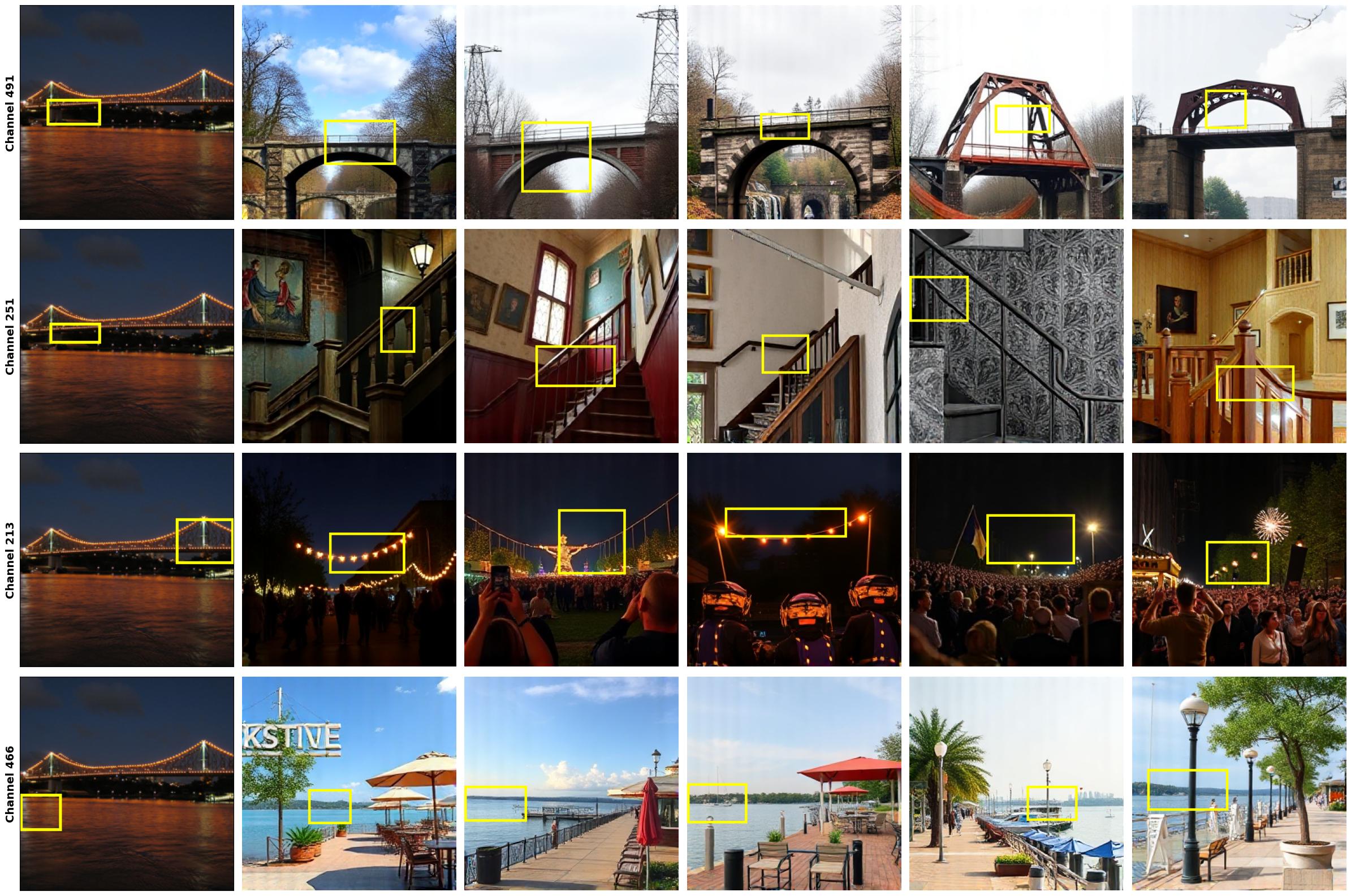}
    \caption{$\mathcal{L}_U$}
\end{subfigure}\hfill
\begin{subfigure}{0.3\linewidth}
    \includegraphics[width=\linewidth, trim=0 0 20cm 0, clip]{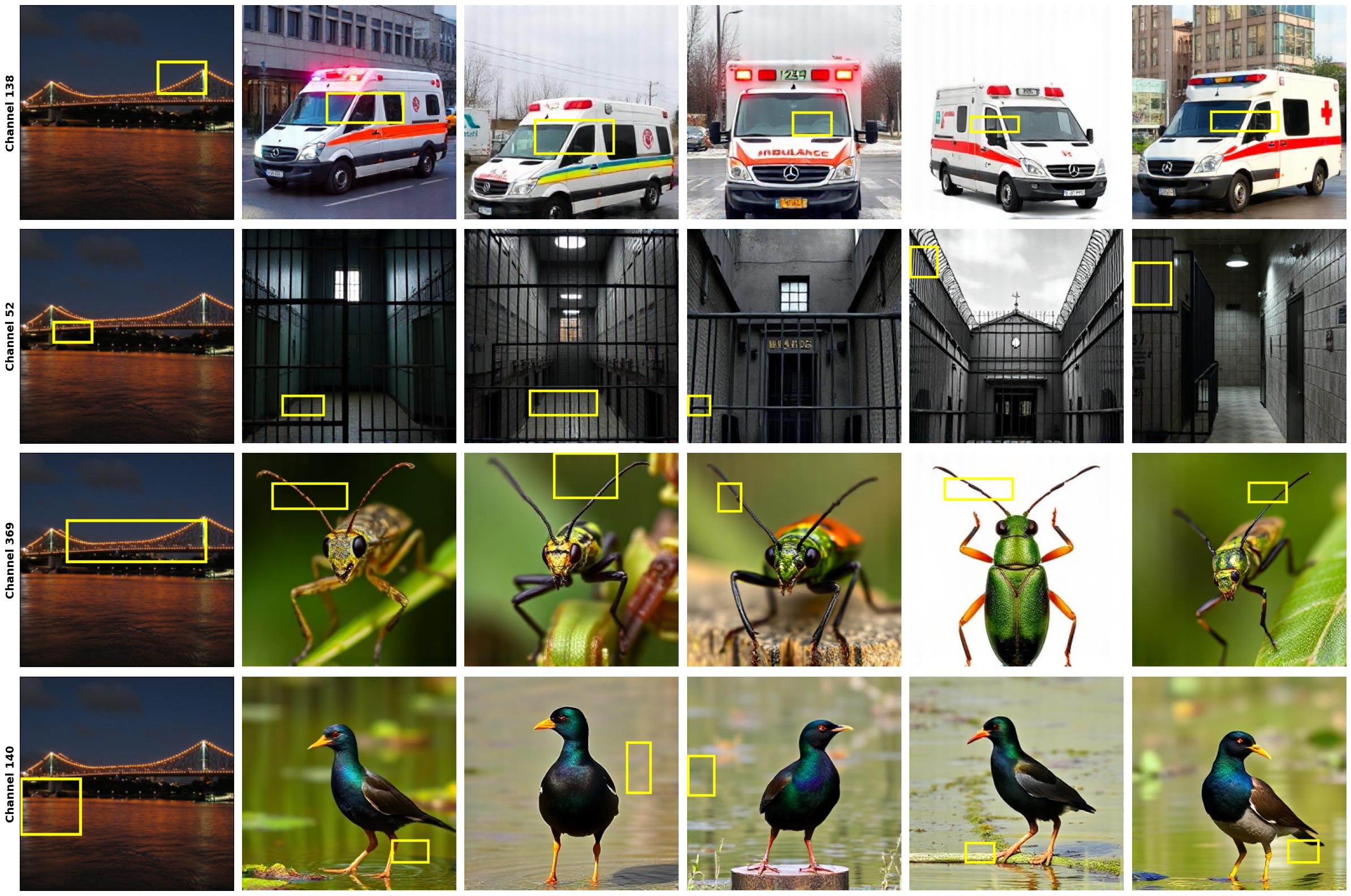}
    \caption{$\mathcal{L}_{\text{reg}}$}
\end{subfigure}

\vspace{0.5em}

\begin{subfigure}{0.3\linewidth}
    \includegraphics[width=\linewidth, trim=0 0 20cm 0, clip]{images/ablation/purity_div/explanation_014.jpg}
    \caption{$\mathcal{L}_U + \mathcal{L}_{\text{div}}$}
\end{subfigure}\hfill
\begin{subfigure}{0.3\linewidth}
    \includegraphics[width=\linewidth, trim=0 0 20cm 0, clip]{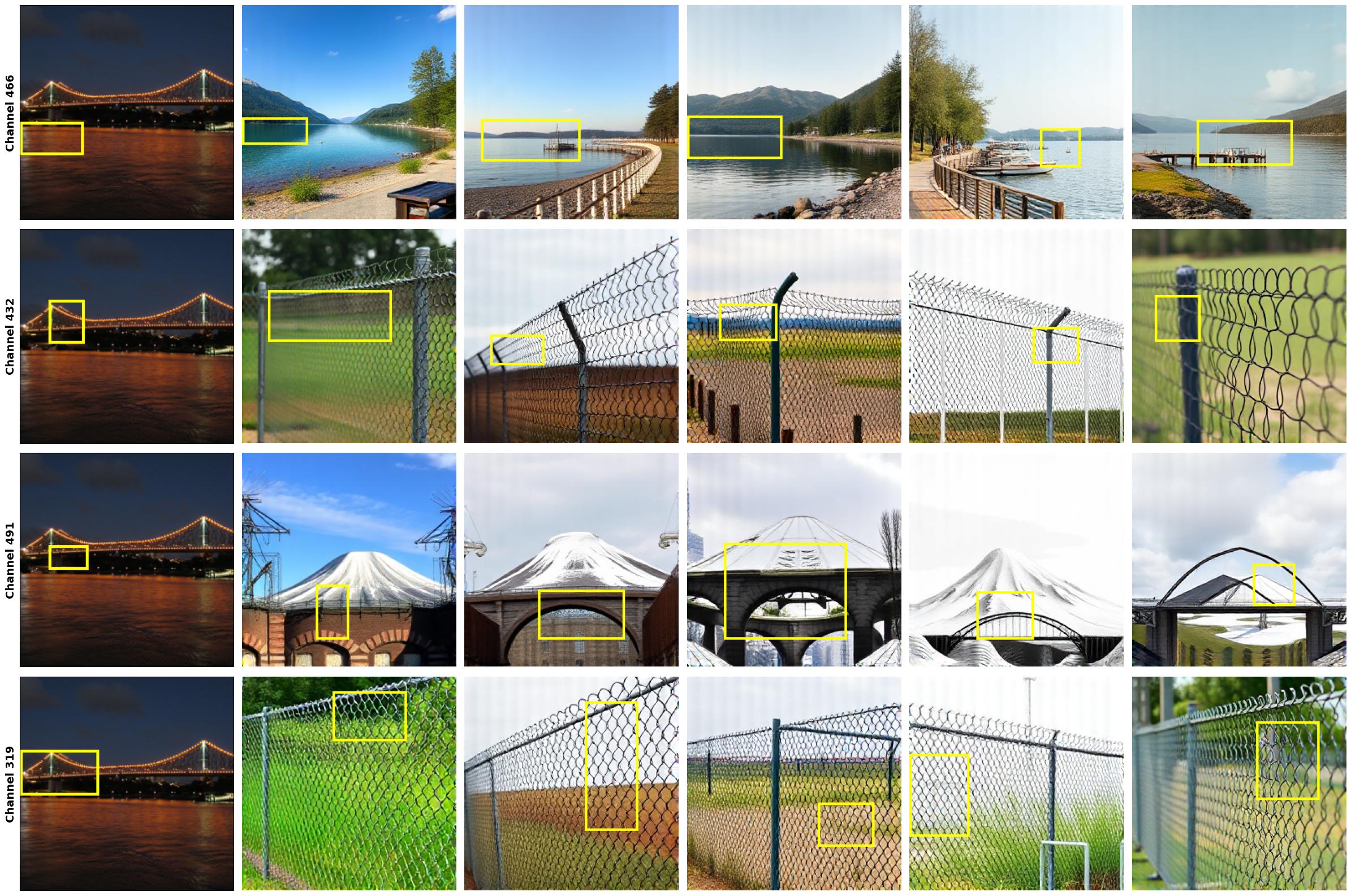}
    \caption{$\mathcal{L}_U + \mathcal{L}_{\text{reg}}$}
\end{subfigure}\hfill
\begin{subfigure}{0.3\linewidth}
    \includegraphics[width=\linewidth, trim=0 0 20cm 0, clip]{images/ablation/reg_div/explanation_014.jpg}
    \caption{$\mathcal{L}_{\text{reg}} + \mathcal{L}_{\text{div}}$}
\end{subfigure}

\vspace{0.5em}

\begin{subfigure}{0.3\linewidth}
    \centering
    \includegraphics[width=\linewidth, trim=0 0 20cm 0, clip]{images/ablation/full/explanation_014.jpg}
    \caption{$\mathcal{L}_U + \mathcal{L}_{\text{reg}} + \mathcal{L}_{\text{div}}$}
\end{subfigure}

\caption{\textbf{Ablation study over loss configurations I.} A qualitative comparison of models trained with different subsets of the objective terms $\{ \mathcal{L}_U, \mathcal{L}_{\text{reg}}, \mathcal{L}_{\text{div}} \}$, including all seven non-empty combinations. Only the full model (bottom) effectively preserves both diversity and structural integrity.}
\label{fig:app_full_ablation}
\end{figure}

This section presents qualitative results for all loss configurations evaluated in our ablation study, as described in the main paper.  Fig.~\ref{fig:app_full_ablation} presents qualitative differences across all loss configurations. 

The results illustrate the role of each component in shaping representation quality. Removing $\mathcal{L}_U$ leads to noticeable semantic drift, while excluding $\mathcal{L}_{\text{div}}$ reduces diversity in learned representations. Configurations without $\mathcal{L}_{\text{reg}}$ exhibit decreased stability and poorer structure in the embedding space. The full objective achieves the most favorable balance between semantic consistency, intra-concept purity, and intra-prototype diversity.

%%%%%%%%%%%%%%%%%%%%%%%%%%%%%%%%%%%%%%%%%%%%%%%%%%%%%%%%%%%%

\begin{figure}[tbh]
\centering

\begin{subfigure}{0.3\linewidth}
    \includegraphics[width=\linewidth, trim=0 0 20cm 0, clip]{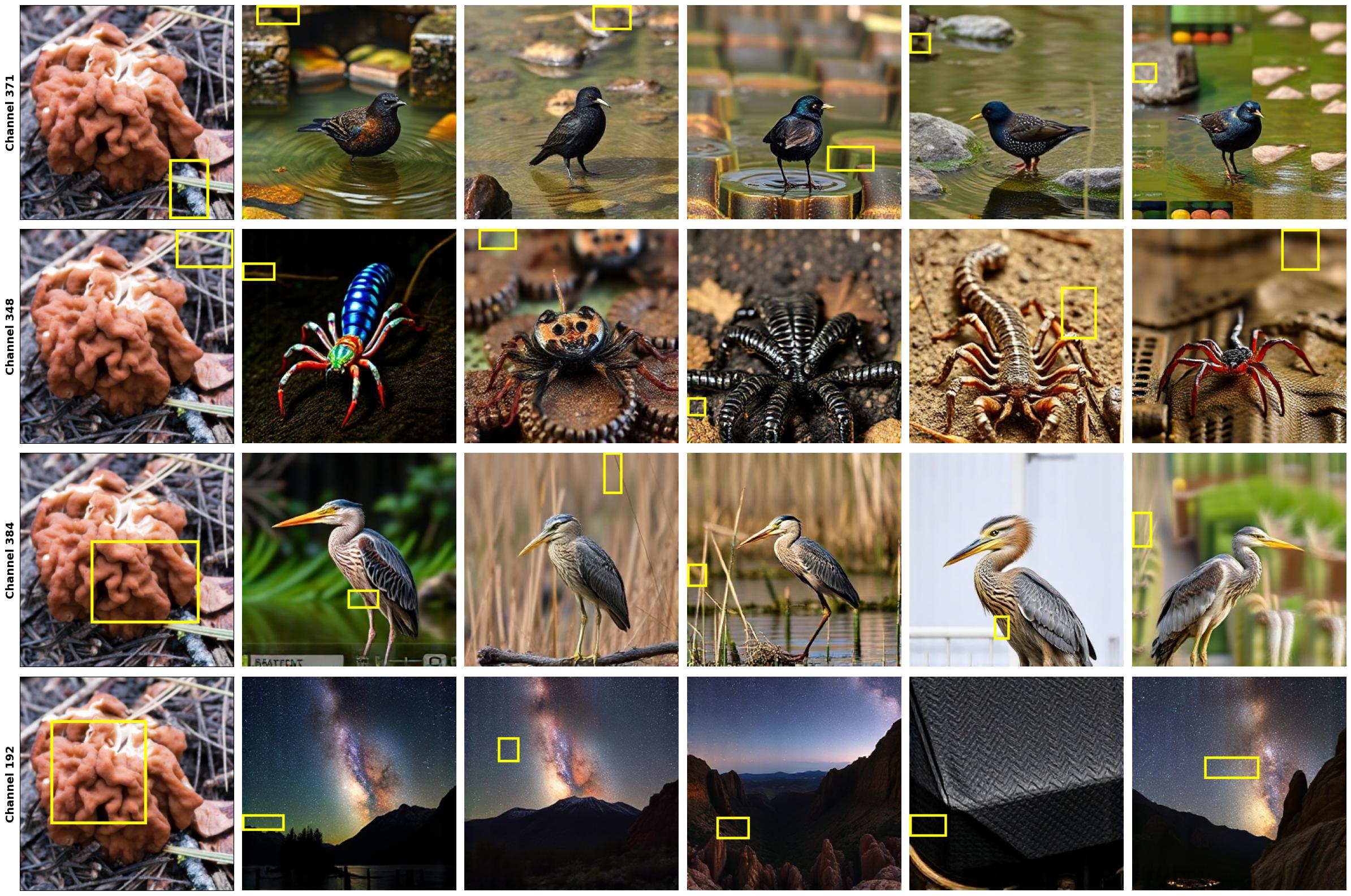}
    \caption{$\mathcal{L}_{\text{div}}$}
\end{subfigure}\hfill
\begin{subfigure}{0.3\linewidth}
    \includegraphics[width=\linewidth, trim=0 0 20cm 0, clip]{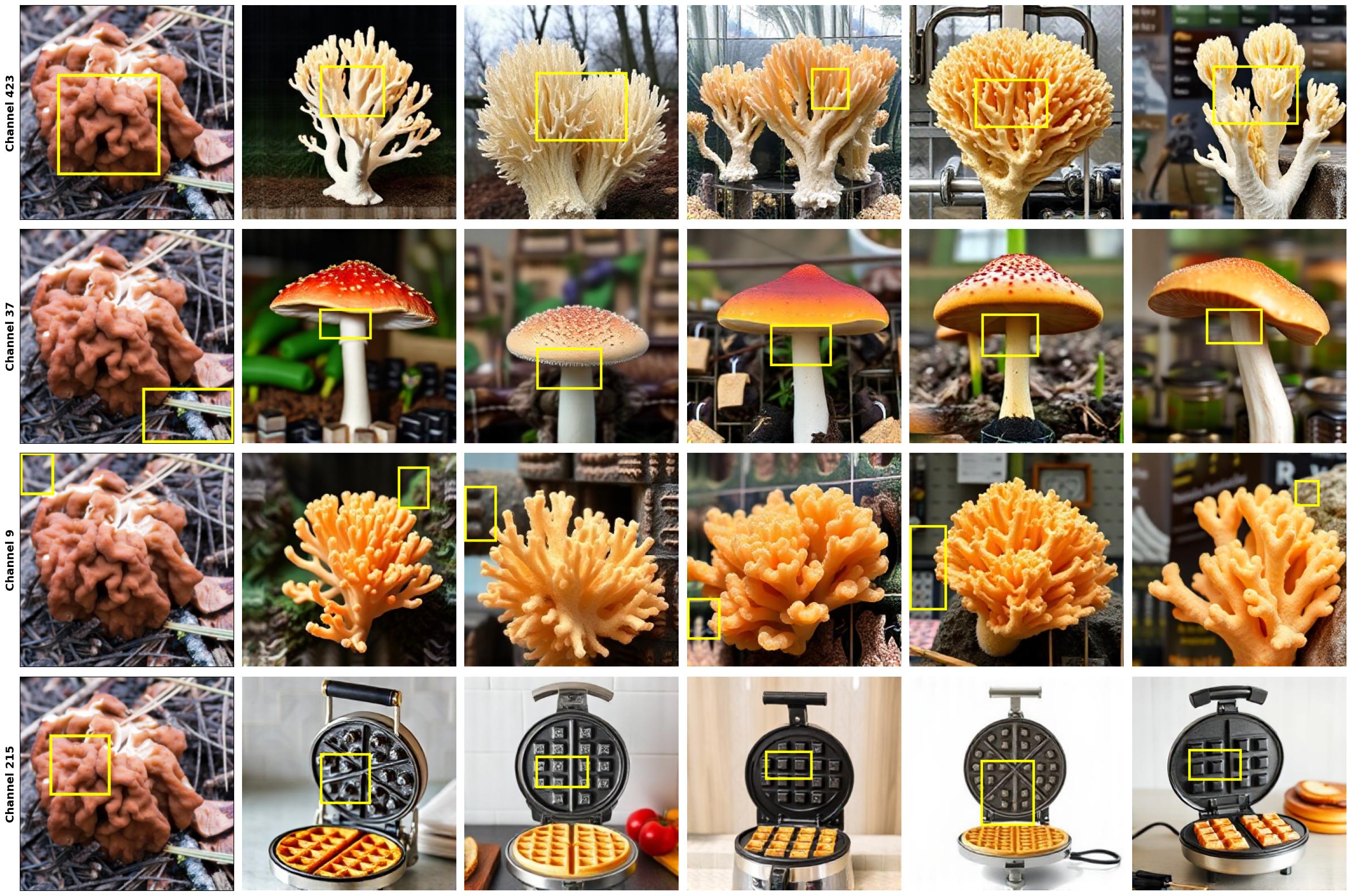}
    \caption{$\mathcal{L}_U$}
\end{subfigure}\hfill
\begin{subfigure}{0.3\linewidth}
    \includegraphics[width=\linewidth, trim=0 0 20cm 0, clip]{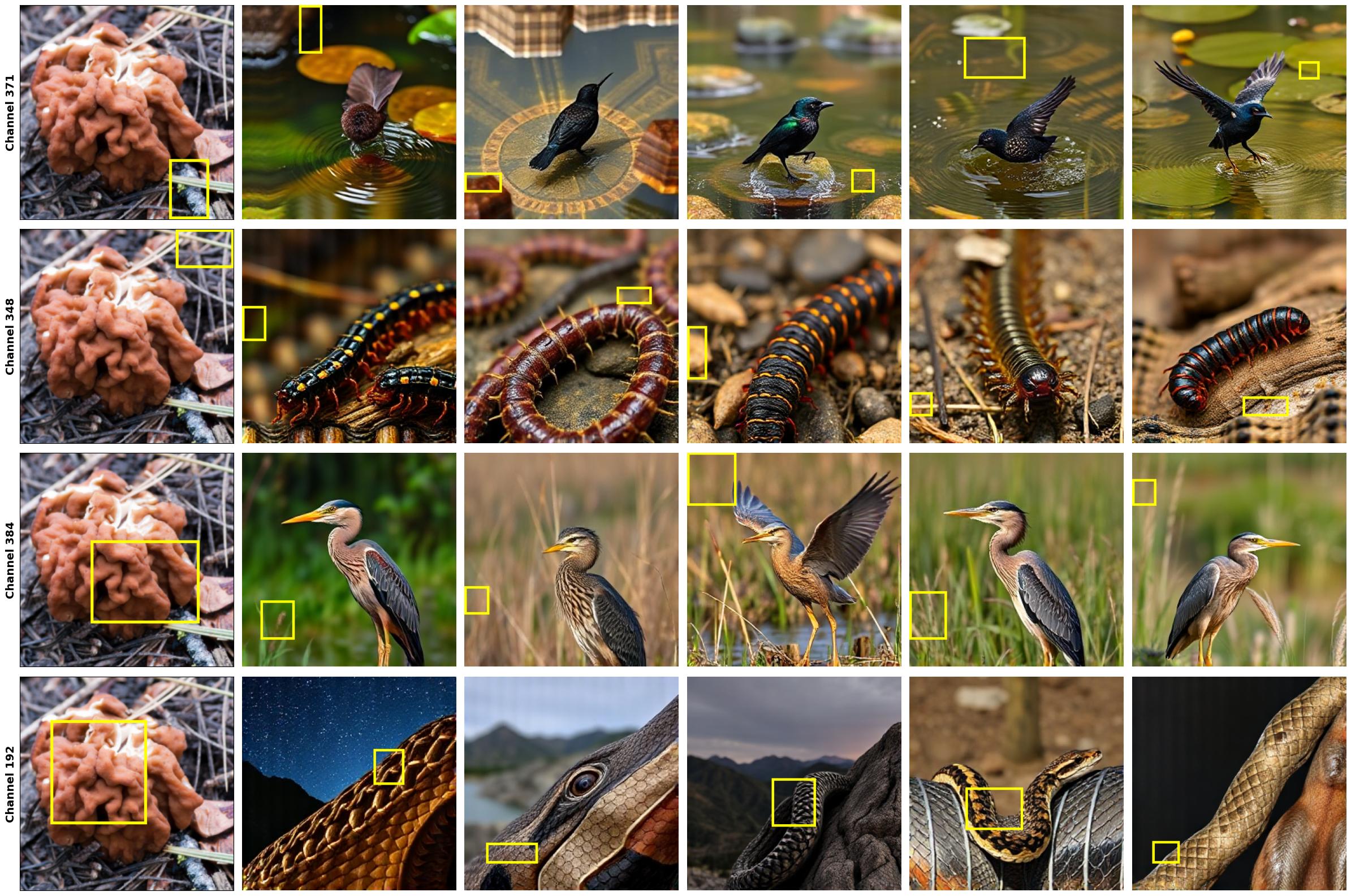}
    \caption{$\mathcal{L}_{\text{reg}}$}
\end{subfigure}

\vspace{0.5em}

\begin{subfigure}{0.3\linewidth}
    \includegraphics[width=\linewidth, trim=0 0 20cm 0, clip]{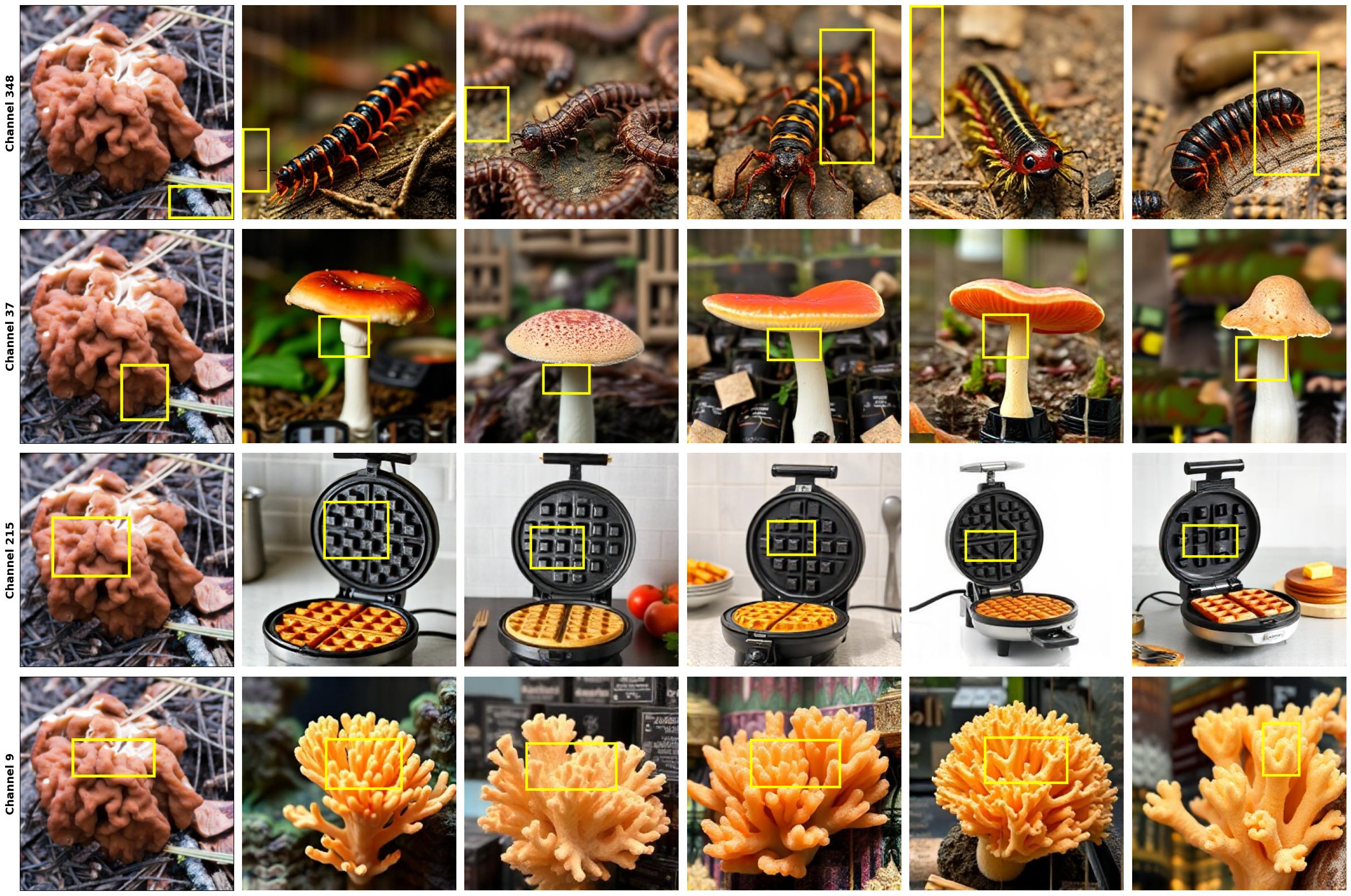}
    \caption{$\mathcal{L}_U + \mathcal{L}_{\text{div}}$}
\end{subfigure}\hfill
\begin{subfigure}{0.3\linewidth}
    \includegraphics[width=\linewidth, trim=0 0 20cm 0, clip]{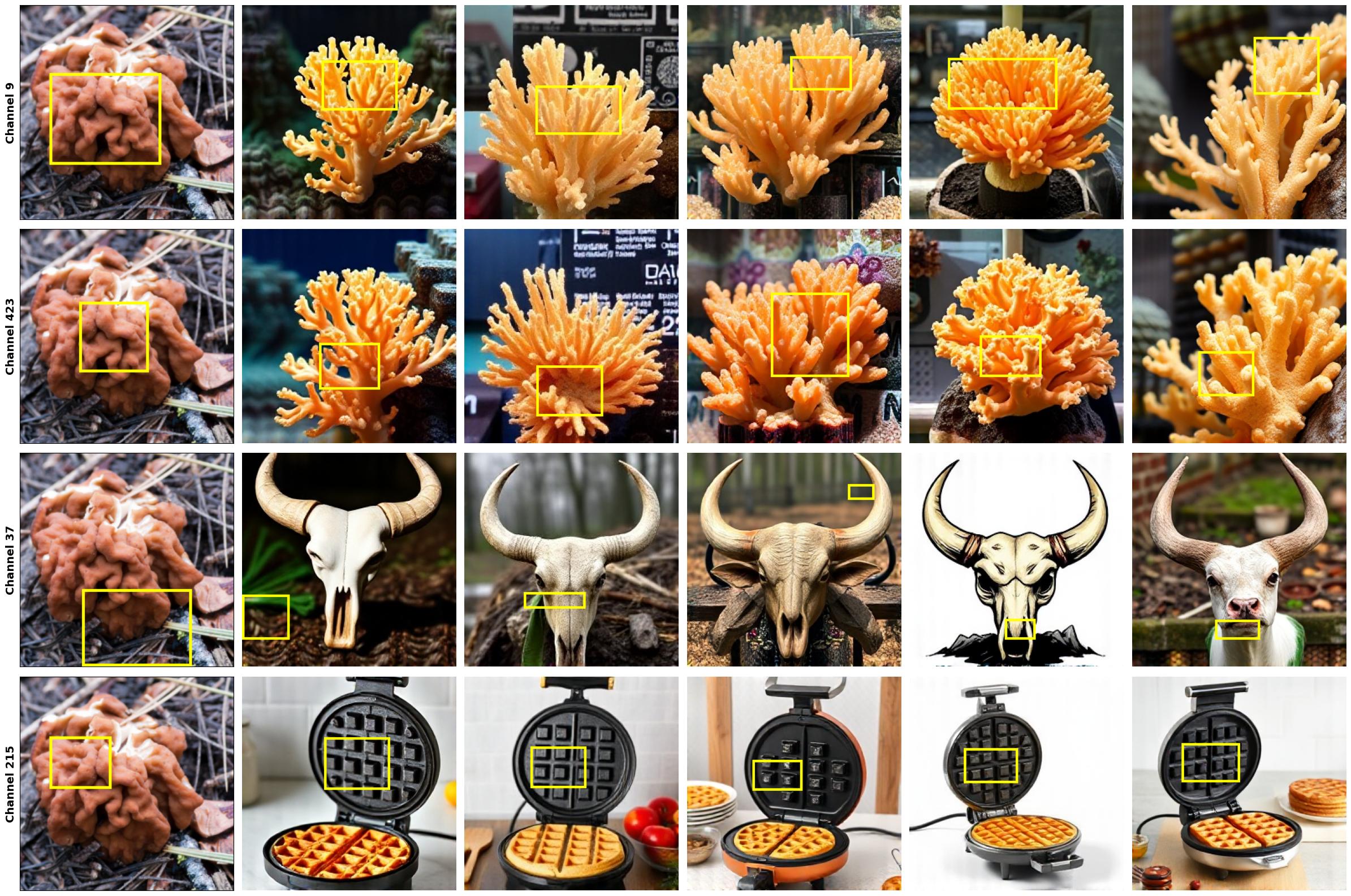}
    \caption{$\mathcal{L}_U + \mathcal{L}_{\text{reg}}$}
\end{subfigure}\hfill
\begin{subfigure}{0.3\linewidth}
    \includegraphics[width=\linewidth, trim=0 0 20cm 0, clip]{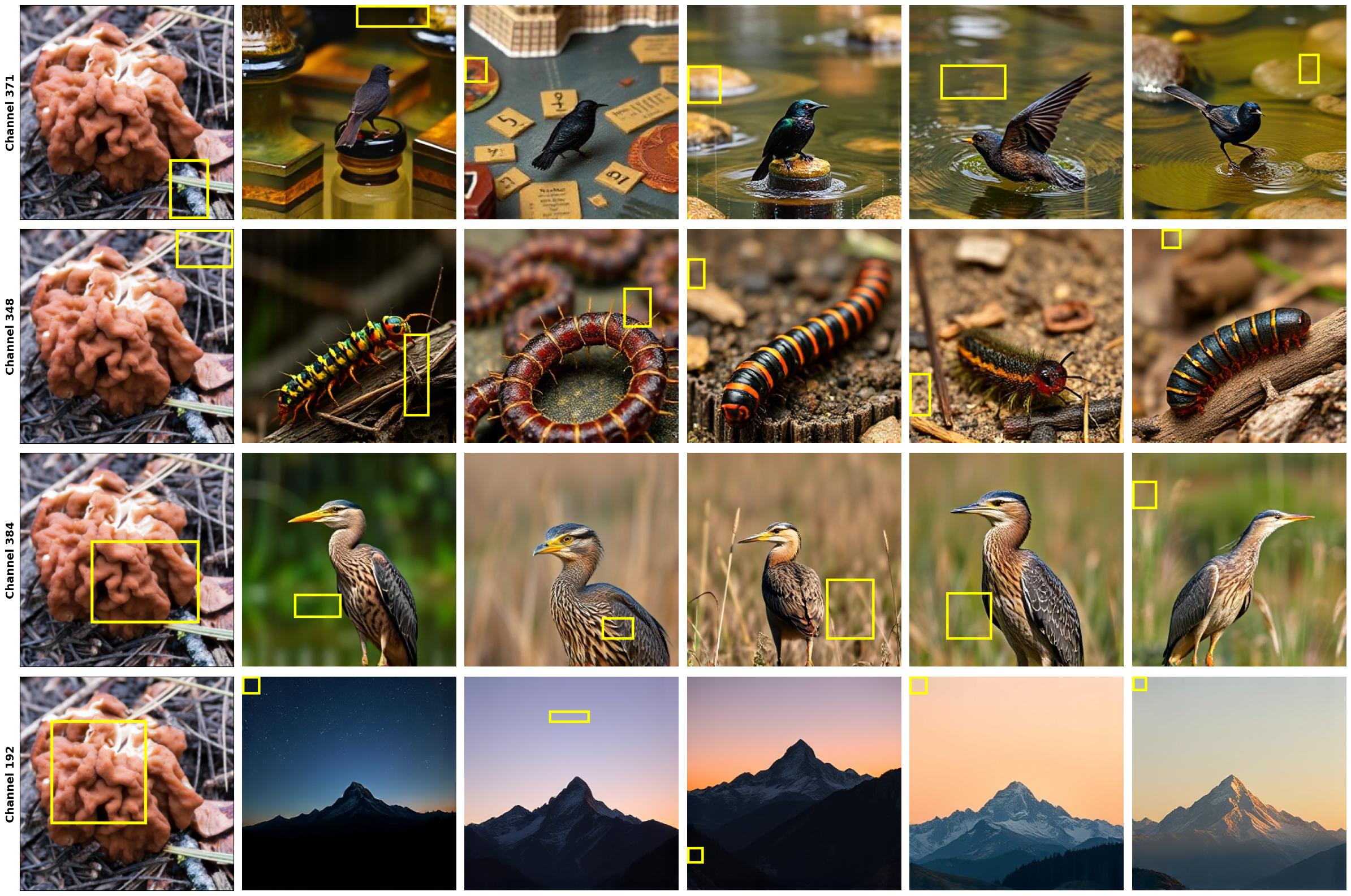}
    \caption{$\mathcal{L}_{\text{reg}} + \mathcal{L}_{\text{div}}$}
\end{subfigure}

\vspace{0.5em}

\begin{subfigure}{0.3\linewidth}
    \centering
    \includegraphics[width=\linewidth, trim=0 0 20cm 0, clip]{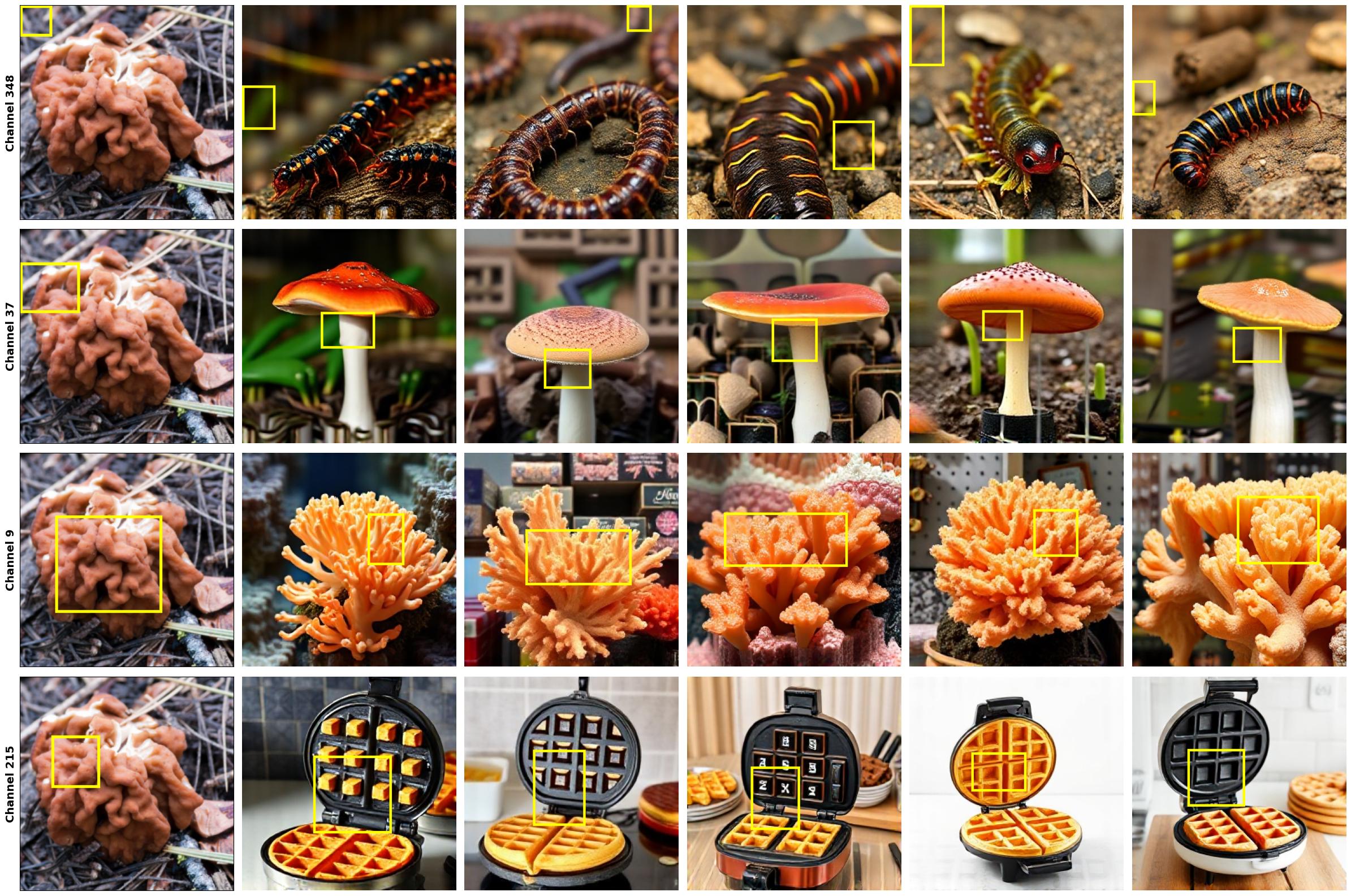}
    \caption{$\mathcal{L}_U + \mathcal{L}_{\text{reg}} + \mathcal{L}_{\text{div}}$}
\end{subfigure}

\caption{\textbf{Ablation study over loss configurations II.} A qualitative comparison of models trained with different subsets of the objective terms $\{ \mathcal{L}_U, \mathcal{L}_{\text{reg}}, \mathcal{L}_{\text{div}} \}$, including all seven non-empty combinations. Only the full model (bottom) effectively preserves both diversity and structural integrity.}
\end{figure}

%%%%%%%%%%%%%%%%%%%%%%%%%%%%%%%%%%%%%%%
% \newpage
% \clearpage
% \input{checklist.tex}

\end{document}